\documentclass[11pt]{article}
\usepackage[T1]{fontenc}
\usepackage[lining]{XCharter}
\usepackage[xcharter,vvarbb]{newtxmath}
\usepackage[type1]{inter}
\usepackage[scale=0.86]{plex-mono}
\usepackage[letterpaper,margin=1.15in,top=1.05in,bottom=1.1in,headsep=0.3in]{geometry}
\usepackage{microtype}
\usepackage{amsmath}

\usepackage{amsthm}
\usepackage{graphicx}
\usepackage{xcolor}
\usepackage{colortbl}
\usepackage{booktabs,multirow,array,tabularx}
\usepackage{tikz}
\usetikzlibrary{arrows.meta,positioning,shapes.geometric,calc,fit}
\usepackage{pgfplots}
\pgfplotsset{compat=1.18}
\usepackage[round]{natbib}
\usepackage{caption}
\usepackage{enumitem}
\usepackage{todonotes}
\usepackage[ruled,vlined]{algorithm2e}
\usepackage{titlesec}
\usepackage{fancyhdr}
\usepackage[colorlinks=true,linkcolor=accentdark,citecolor=accentdark,urlcolor=accentdark]{hyperref}
\hypersetup{pdftitle={Bellman Language Models: Solving the Recurring Decisions of LLM Agents},
  pdfauthor={Adarsh Vatsa, William Eiers}}
\definecolor{dtdline}{HTML}{2563EB}
\definecolor{dtdfill}{HTML}{60A5FA}
\definecolor{llmline}{HTML}{525252}
\definecolor{llmfill}{HTML}{A3A3A3}
\definecolor{hiline}{HTML}{DC2626}
\definecolor{ink}{HTML}{262626}
\definecolor{accent}{HTML}{2563EB}
\definecolor{accentdark}{HTML}{1D4ED8}
\definecolor{muted}{HTML}{737373}
\definecolor{rule}{HTML}{E5E5E5}
\definecolor{tablerule}{HTML}{737373}
\arrayrulecolor{tablerule}

\titleformat{\section}{\sffamily\fontsize{15}{19}\selectfont\bfseries\color{ink}}{\color{accent}\thesection}{0.7em}{}
\titleformat{\subsection}{\sffamily\fontsize{11.5}{15}\selectfont\bfseries\color{ink}}{\color{accent}\thesubsection}{0.65em}{}
\titleformat{\subsubsection}{\sffamily\normalsize\bfseries\color{ink}}{\color{accent}\thesubsubsection}{0.6em}{}
\titleformat{\paragraph}[runin]{\sffamily\normalsize\bfseries\color{ink}}{}{0pt}{}[.]
\titlespacing*{\section}{0pt}{2.2ex plus .6ex minus .2ex}{1.0ex}
\titlespacing*{\subsection}{0pt}{1.8ex plus .5ex minus .2ex}{0.7ex}
\DeclareCaptionLabelFormat{blm}{{\sffamily\bfseries\color{accentdark}#1~#2}}
\DontPrintSemicolon
\SetKwComment{tcp}{\textcolor{muted}{// }}{}

\SetCommentSty{blmcommentfont}
\SetAlCapFnt{\small\sffamily\bfseries\color{accentdark}}
\SetAlCapNameFnt{\small}
\newtheoremstyle{blmplain}{0.6em}{0.6em}{\itshape}{}{\sffamily\bfseries\color{accentdark}}{.}{0.5em}{}
\theoremstyle{blmplain}
\newtheorem{proposition}{Proposition}
\fancypagestyle{plain}{\fancyhf{}}

\newcommand{\blm}{BLM}
\newcommand{\tautwo}{$\tau^2$}

\makeatletter
\newcommand{\subtitle}[1]{\gdef\@subtitle{#1}}
\newcommand{\affiliation}[1]{\gdef\@affiliation{#1}}
\renewcommand{\maketitle}{%
  \thispagestyle{plain}%
  \begingroup\raggedright
  {\sffamily\fontsize{8}{10}\selectfont\color{accentdark}\bfseries PREPRINT \textcolor{muted}{\textperiodcentered{} \mdseries October 2026}\par}
  \vspace{10pt}
  {\sffamily\fontsize{26}{30}\selectfont\bfseries\color{ink}\@title\par}
  \vspace{6pt}
  {\sffamily\fontsize{14}{18}\selectfont\color{muted}\@subtitle\par}
  \vspace{14pt}
  {\sffamily\fontsize{10.5}{13}\selectfont\color{ink}\bfseries\@author\par}
  \vspace{2pt}
  {\sffamily\fontsize{9}{11}\selectfont\color{muted}\@affiliation\par}
  \endgroup
  \vspace{12pt}}
\renewenvironment{abstract}{%
  {\color{rule}\hrule height 0.6pt}\vspace{8pt}%
  {\sffamily\fontsize{8}{10}\selectfont\bfseries\color{accentdark}ABSTRACT\par}\vspace{3pt}%
  \small}{\par\vspace{8pt}{\color{rule}\hrule height 0.6pt}\vspace{4pt}}
\makeatother

\title{Bellman Language Models}
\subtitle{Solving the Recurring Decisions of LLM Agents}
\author{Adarsh Vatsa\quad William Eiers}
\affiliation{Stevens Institute of Technology}
\date{}

\begin{document}
\maketitle

\begin{abstract}
A general observation in agentic workflows, especially bounded ones, is that the decision space of the LLM, when we separate a decision from generation, starts approaching saturation: the agent meets the same states again and again, and after enough conversations it stops meeting new ones. LLMs generally don't treat these decisions as an optimization problem and instead re-derive them from their context/prompt on every turn. We show that treating decision making within reasonable bounds as an optimization problem, solving it once, and reusing the solved decisions can often make an agent faster, cheaper and more reliable. Our approach, which we call Discretize, Then Decide, starts from data such as the workflow's own logs or, when none exists, runs of the LLM agent itself. It reads each conversation into discrete states and estimates the probability of each outcome and the cost of each move, until no new states appear and the probabilities stop changing. It then solves the resulting Markov decision process. The solution runs as a controller around the model, sending moves whose content is fixed in the state, such as a tool call or a templated reply, and any state it hasn't seen is sent back to the LLM. We call an LLM run this way a \emph{Bellman language model} (BLM), after Richard Bellman's equation for the best move in each state.

On \tautwo-bench telecom, the controller with no model solves 79\% of tasks. As a BLM, Qwen3.8-27B, an open model, goes from 29\% to 96\% with 93\% fewer model calls. For Claude Opus 5.5, which solves 96\% of tasks on its own, running it as a BLM cuts model calls and cost per task by about 98\% and lets it solve 2 more tasks. On InventoryBench, the solved policy with no model beats every published method. Where the hard part is writing content rather than choosing moves, such as resolving SymPy issues, few decisions can be solved in advance, and a BLM helps little.
\end{abstract}

\section{Introduction}\label{sec:intro}

A general observation about LLM agents in bounded workflows is that, once we separate a decision from the generated text used to carry it out, the space of decisions saturates. The agent meets the same states again and again, and after enough
conversations it stops meeting new ones.

For example, take an agent on \tautwo-bench's telecom domain \citep{barres2025tau2} helping a customer who can't send picture messages.
A \emph{fault} is something wrong with the phone or the line that stops picture messages, such as airplane mode left on. A \emph{check} is a question or instruction to the customer that finds out whether one fault is there, such as asking them to look for the airplane icon.

The agent's replies differ in every conversation, but its decision at each turn is one of a short list. It can look up the account, run one of its checks, fix a fault, take a payment, hand the customer to a human, or close. That's what we mean by a bounded workflow: its moves can be listed.

To choose its next move, the agent needs only a handful of facts. What's the problem, and is it fixed yet? Is anything
stopping a fix, such as a locked SIM card? Which checks are left to try? We call each combination of these facts a state.

The replies keep changing, but the states don't. Almost every conversation we collected has messages the agent never sent
before, yet on new tasks it almost never meets a state it hasn't met before (Figure~\ref{fig:saturation}). One might think that just means the states keep too few facts. But in telecom they keep enough to decide well: a controller deciding from these facts alone, with no model, completes most tasks (Section~\ref{sec:results}). So the states have to keep exactly what the decision depends on.

Yet LLM agents generally don't treat these decisions as an optimization problem. They re-derive each one from their context
on every turn, with a new call to the model, so they pay for the same decisions over and over.

They also don't always decide well. Claude Opus 5.5 \citep{opus55} tends to ask about faults that never occur before the
most common one. Qwen3.8-27B \citep{qwen38}, an open-weights model, mostly fails by handing the customer to a human while
there are still checks to try. This isn't for lack of information. In a simulated coding task where we know the true probabilities, we asked an LLM to estimate them and solved the optimization problem with its estimates. The decisions were
as good as with the true probabilities, while the LLM choosing for itself did worse (Section~\ref{sec:results}).

Decisions that recur like these are what operations research learned to solve. Ordering stock, for example, became a
problem with a stated cost of ordering and a stated cost of running out, solved once and applied to every order
\citep{arrow1951optimal}. Dynamic programming did the same for sequences of decisions \citep{bellman1957dp}. Earlier
dialogue systems brought this approach to conversations, planning each move with a partially observable Markov decision
process \citep{williams2007pomdp,young2013pomdp}. But they understood users only through narrow, error-prone components,
and mostly replied from templates. LLMs do both well.

So solving recurring decisions once isn't new. What, then, is the contribution of this paper? We show that the idea holds
up for LLM agents, and we make it work with them. For people who build agents, that's a recipe that starts from the
agent's own logs (Section~\ref{sec:prescription}). For people who study them, it's evidence that their decisions recur,
and that they don't optimize them.

\begin{figure}[!t]
\centering
\includegraphics{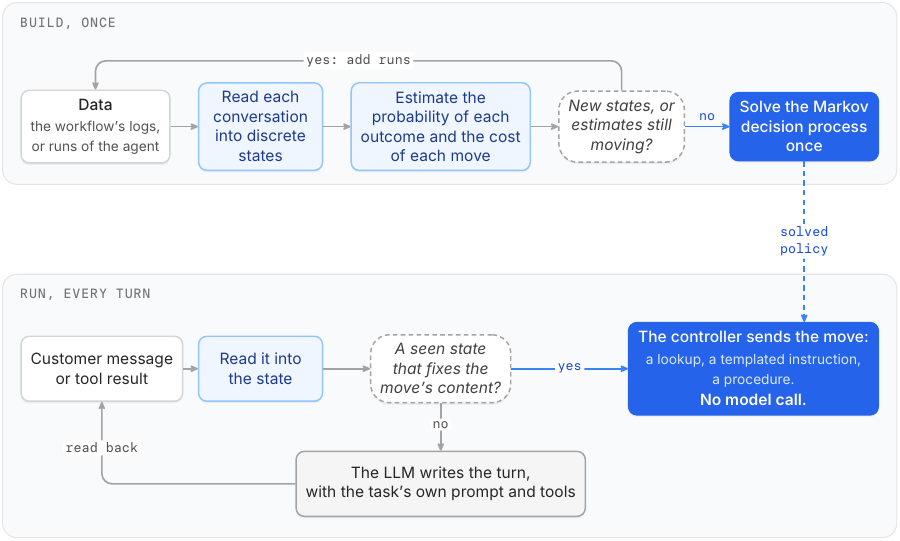}

\medskip
\includegraphics{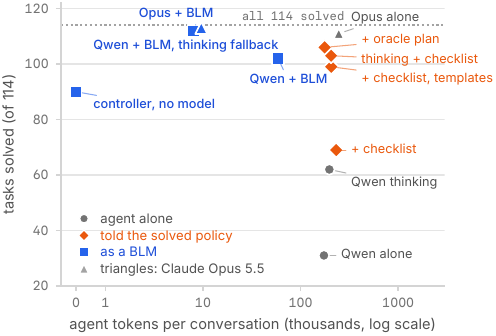}
\caption{Discretize, Then Decide. Top: built once, run every turn. The key point is that when the state fixes the move, the turn never reaches the model. Bottom: tasks solved against agent tokens in telecom, where up and to the left is better. Telling the model the solved policy moves it up. Executing it moves it up and left. Other variants: Table~\ref{tab:ablate}.}
\label{fig:method}\label{fig:pareto}
\end{figure}

\emph{The key idea is to treat an agent's recurring decisions as an optimization problem: solve it once from data, then
reuse the solution instead of re-deriving each decision on every turn.} We call this method Discretize, Then Decide. It
starts from the workflow's logs, or runs of the LLM agent. As the top of Figure~\ref{fig:method} shows, it reads
each conversation into discrete states with simple rules, such as regular expressions over messages and tool
results. In practice, the rules can be simple, because they only look for the handful of facts the decisions depend on. Whatever they can't
read goes to the LLM.

It then estimates, for each move, the probability of each outcome and the move's cost, adding runs until no new states
appear and the estimates stop changing. It solves the resulting Markov decision process for the lowest expected cost. The solution, the \emph{solved policy}, says which move to make in each state. Whoever runs the agent
chooses the unit of cost, such as customer turns in a support conversation. Every move
and every outcome has a price in that unit.

The solution runs as a controller around the model. The controller reads every message and tool result into the state.
When the state fixes what to send, such as an account lookup or a templated instruction, the controller sends it.

Any other turn, including every turn in a state it hasn't seen, goes to the LLM with the task's prompt and the conversation so far. Crucially, the model is never called on the turns the solved policy decides. We call an LLM run this way a \emph{Bellman language model} (BLM), after Richard Bellman's equation for the best move in each state \citep{bellman1957dp}. Discretize, Then Decide is how we build one.

This differs from the usual ways of making agents cheaper or more reliable. Routers and cascades choose which model answers
each query \citep{chen2023frugalgpt,ong2024routellm}, while a \blm{} decides whether a turn needs a model at all. Couldn't we just
write the solved policy into the prompt? Then the model still decides every turn \citep{xiao2024flowbench}. In our
experiments that makes the model more accurate than on its own, but no more efficient. Executing the solved policy does
better on both (bottom of Figure~\ref{fig:pareto}).

Other systems carry out or enforce a written policy
\citep{bocklisch2024calm,rebedea2023nemo,policyguide2026,nomos2026}, or have a model write the agent's program, or compile
its tasks or past runs into code \citep{qi2026abcagent,agentjitcomp2026,agentjitgithub}. Either way, a person or a model
wrote the decisions. In our experiments a coding agent wrote each \blm's rules, templates and solver too. It worked from
the workflow's guide, tools and logs (Section~\ref{sec:implementation}). But no one writes its decisions: the solver, a standard algorithm such as dynamic programming, picks the best move from estimated probabilities and stated costs, as Section~\ref{sec:related} explains.

We evaluate \blm{}s in four domains, chosen to range from nearly all decisions to nearly all content. They are troubleshooting
customers' phones in \tautwo-bench's telecom domain, ordering stock under uncertain demand and delivery in InventoryBench
\citep{inventorybench2026}, triaging failed test jobs in Mozilla's continuous integration (CI), and resolving SymPy issues. The
LLMs are Qwen3.8-27B and Claude Opus 5.5.

In telecom we built the controller three times, each time fixing what the last test showed. We built each from training
tasks only, froze it, and then drew its own test tasks at random from the rest. We report them separately, and
Appendix~\ref{app:ledger} lists every test, each recorded publicly before any result.

In telecom, the controller with no model at all solves most tasks. On a turn where its state doesn't fix the move, it just sends one fixed request and reads the reply. With Qwen3.8-27B writing those turns, it solves about as many tasks as Claude Opus 5.5 alone, with
far fewer model calls. Fine-tuning Qwen3.8-27B on Opus's own conversations instead makes it solve fewer tasks than the \blm{} does, and it still calls the model on every turn. With Opus writing those turns, the \blm{} removes almost all model calls, cost and waiting time, and solves slightly more (bottom of Figure~\ref{fig:pareto}).

On one set of tasks, each with a fault that never appeared in the training data, we ran every task three times.
As a Bellman model, Qwen3.8-27B solves nearly all of them on all three attempts, against fewer than half on its own.

How does it handle a fault it never saw? The checks come from the workflow's own guide, not from the data. So the check for a new fault is still on the list, just last, because its fault never occurred in training. The state isn't new either. It holds the kind of problem, whether it's fixed yet and which checks are left, not the fault.

When the whole task is deciding, the solved policy needs little or no model. In inventory control it needs none, and it beats every published method. In test triage it pays an LLM only to read some logs, and the evidence is weaker. There it costs about as much as a fixed rule that reruns a failed test once and then escalates.

Where the hard part is writing content rather than choosing moves, as in resolving SymPy issues, few of the agent's turns are decisions a state can fix. So a \blm{} helps little. The agent's own logs show this before anyone builds a \blm. Replaying a draft controller over them
shows how many of the agent's turns its state would fix. That's nearly all of Opus's turns in telecom, against a small share
of the SymPy agent's turns.

The paper makes four claims:
\begin{enumerate}[leftmargin=*,itemsep=2pt,topsep=2pt]
\item \textbf{Saturation.} Once we separate deciding from writing, a bounded workflow's states saturate, and the
text doesn't. We show this in telecom (Observation~1).
\item \textbf{Optimality.} LLM agents generally don't optimize their decisions: a solver given the probabilities they state decides better
(Observation~2).
\item \textbf{Execution.} Executing the solved decisions removes the model calls they would cost. Where most of the work is
deciding, that removes most of the agent's cost and waiting time, and its accuracy and consistency hold or improve. The gain
comes from executing, not from telling the model (Observations~3 to 7).
\item \textbf{Prediction.} The agent's own logs predict where this pays (Observation~8).
\end{enumerate}

If these claims hold, an agent doesn't have to pay again for states it has met before. Our strongest evidence comes from telecom, where LLMs play the customers, who may vary less than real ones. The other three domains show how far the
claims carry, but there we kept test data apart from development less strictly (Section~\ref{sec:protocol}). Choosing the states is itself an optimization problem, and it isn't the one we solve. When we call a decision optimal, we mean optimal for the states we chose, and Section~\ref{sec:limits} leaves the best choice of states open.

\section{Preliminaries}\label{sec:prelim}

This section sets out the agent's world in simple terms: first in general, then in telecom, and then the decision problem they share.

In general, an LLM agent in a bounded workflow works through \emph{tasks}. Each task is one instance of the same job. For example, it can be a support case, a failed test job, a stretch of weeks of ordering stock, or an issue to fix in a code base. On each \emph{turn} it
makes one \emph{move} from a short list. Some moves gather information, such as a lookup or a question. Some act, such as a
fix, an order or an edit. And some end the task, such as closing it or handing it to a person.

Some of what matters stays hidden from the agent, such as which faults a phone has, or whether a failed test is a real regression. The workflow usually comes with written rules about what the agent may do, and every move has a cost, in one unit that whoever deploys the
agent cares about.

Concretely, in telecom, each task is one support case. A customer can't do something with their phone, such as send picture
messages, because something on the phone or the line is wrong. Each such problem is a \emph{fault}: airplane mode left
on, roaming off while the customer is abroad, or the network set to 2G only. A case can have several faults at once, and
it's solved only when all of them are fixed.

The agent and the customer take turns. A \emph{check} asks the customer to look at one setting, such as whether airplane mode is on, to find out whether its fault is there. If it is, the check also asks them to change the setting. Some faults also need a \emph{fix} on the account, made with a \emph{tool call}, such as enabling roaming on the line. The agent can also \emph{look up} the account, the line or the bill, close the conversation, or \emph{hand the customer to a
human}, which we call a \emph{hand-off}.

In telecom, the benchmark gives the agent two documents. The \emph{troubleshooting guide} lists the checks in three \emph{stages}: the line's
service, then mobile data, then picture messaging. The \emph{written policy} sets rules the agent has to follow, such as allowing a hand-off when the SIM card is locked. Some of its rules are \emph{procedures}, fixed series of steps, such as taking a payment before resuming a suspended line.

Telecom has the shape of troubleshooting: the hidden facts are faults, the information moves are checks, and each check also
fixes what it finds. Other workflows fill the same parts differently.

In continuous integration, the hidden fact is whether a failed test job is a real regression or an intermittent failure. The information moves are rerunning the job and paying an LLM to read its error lines. Each task ends by marking the failure intermittent, backing out the change, or escalating to a person.

In inventory control there's nothing to ask: news about demand and deliveries arrives on its own, and the only move is how much to order. In SymPy, the hidden fact is which phase of its work the coding agent is in, and the information moves are searching, reading and running code. Most of the work is the agent's own edits. Table~\ref{tab:domains}, in Appendix~\ref{app:domains}, gives each domain's parts.

Which move should come next? That's a decision problem with three parts: the \emph{states} a task can be in, the moves
available in each, and what each move costs and leads to. In telecom, a state holds the four facts from the introduction: the kind of problem, whether it's fixed yet, anything stopping a fix, and which checks are left.

A \emph{policy} says which move to make in each state. The
\emph{solved policy} is the best one we can compute from the estimates, with the lowest expected total cost. We compute it once, before any conversation. A \emph{controller} then sends its moves, as tool calls or \emph{templated messages}, and the LLM
writes everything else.

So a \blm{} saves in two ways. What the solved policy minimizes depends on the workflow: customer turns in telecom, dollars in
CI, tokens in SymPy. And every move the controller sends costs no model call at all, whatever the solved policy minimizes.
That second saving is the one our experiments show most clearly: in telecom, the \blm{}s removed 93 to 98\% of the model calls
(Section~\ref{sec:results}). We build each of these up step by step in Section~\ref{sec:method}, and write them down formally in Section~\ref{sec:problem}.

It's tempting to read ``bounded'' as ``small''. It isn't. A workflow is bounded when we can list its moves and read the facts its decisions depend on, however many there are. Picture an airline rebooking thousands of passengers after a storm, or a hospital assigning beds and staff through the night. Or picture a warehouse deciding what to pick and restock, or a cloud provider working incidents across thousands of services. Each has a finite menu of moves, facts we can read, and decisions that come back again and again.

Their state spaces are huge, but solving large decision problems is what modern compute does well, and we have to do it only once. So we can generally put decision making in bounded environments into this shape, and leave the LLM to handle the language and whatever the controller hasn't seen before. What limits it isn't the
size of the environment, but whether its states repeat often enough and whether we can observe its outcomes (Section~\ref{sec:theory}).

\section{Discretize, Then Decide}\label{sec:method}

How can an agent stop re-deriving decisions it has met before? Read its conversations into discrete states, and solve the
resulting decision problem once. A controller can then make those decisions on every turn without calling the model, and
send every state it hasn't seen back to the LLM.

This section builds a Bellman model on a simpler version of one task, one step at a time. Then we state the decision problem in
general, give the algorithm, and show what changes for the real telecom controller.

\subsection{A worked example}\label{sec:toy}\label{sec:example}

Take the customer from the introduction, who can't send picture messages. In our worked example, only three things can be wrong, alone
or together. Concretely, the customer may be abroad with roaming off on the line, the phone may be in airplane mode, or its
network mode preference may be set to 2G only. For each fault the agent has one check, which finds the fault and fixes it. The agent
can also hand the customer to a human, or close once the customer says picture messages work.

These three faults come from a real task in our test set, which Section~\ref{sec:build} returns to. The real workflow has lookups and 14 checks, which the troubleshooting guide sorts into three stages. We keep just enough in the example to follow each step by hand.

\subsection{Step 1: read each conversation into discrete states}\label{sec:step-read}

An agent's replies differ in every conversation, but what its next move depends on is small. In the example, it's which checks
the agent has made, what the customer reported, and whether picture messages work yet. We call each combination of these facts
a state. A \blm{} reads every message into the state with simple rules, which we call the reader. It reads the agent's own
messages too, to know which checks the agent has made.

\begin{figure}[t]
\centering
\includegraphics{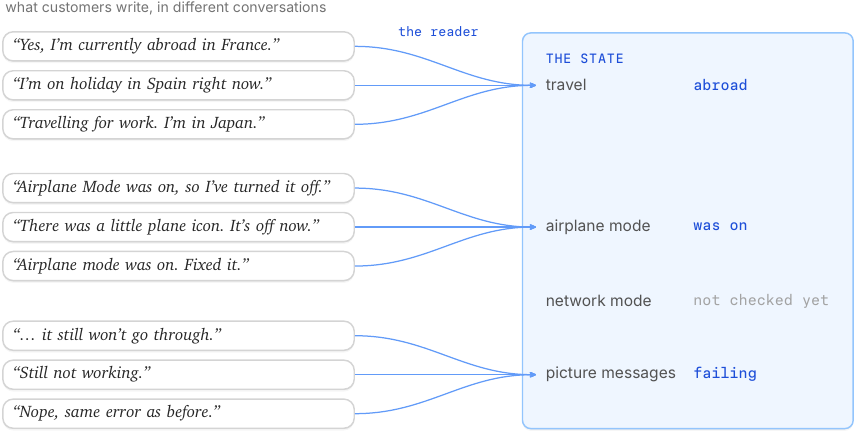}
\caption{Step 1 on the worked example: reading conversations into states. Customers word the same fact in many ways. Notice that the reader
keeps only the fact, so different conversations end up in the same state.}
\label{fig:read}
\end{figure}

As Figure~\ref{fig:read} shows, customers say they're abroad in many ways, but the reader keeps only the fact that they're
abroad. The wording is gone, so different conversations end up in the same state. However a conversation goes, it's
always in one of a few states, and the agent meets the same states again and again.

\subsection{Step 2: estimate the probability of each outcome and the cost of each move}\label{sec:step-estimate}

\begin{figure}[t]
\centering
\includegraphics{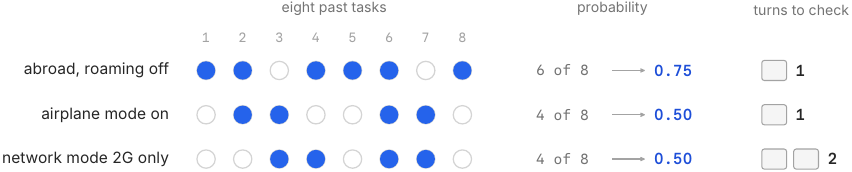}
\caption{Step 2 on the worked example: estimating. Each column is a past task, with a dot for each fault it had. Counting the dots gives
the probability of each fault. Notice that roaming is off in six of the eight, the most common fault.}
\label{fig:estimate}
\end{figure}

Next, we count. For a check, the outcome that matters is whether it finds a fault. Figure~\ref{fig:estimate} shows eight
past tasks, each with the faults it really had, much as a support center's closed tickets record what was wrong. For example, roaming was off for a customer abroad in six of the eight, so the probability of that fault is 0.75. Airplane mode and the 2G-only
setting each appear in four, so theirs are 0.5.

The cost of a check is the number of customer turns it takes. In the example, asking about travel takes one turn, and so does
checking airplane mode. The network mode takes two. In the real workflow we measure turns in logged conversations, and we
keep adding conversations until no new states appear and the probabilities stop changing.

\subsection{Step 3: solve once, before any conversation}\label{sec:step-solve}

\begin{figure}[t]
\centering
\includegraphics{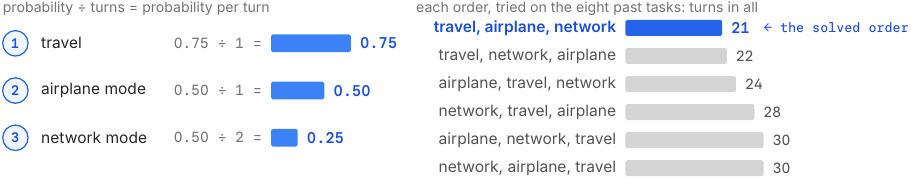}
\caption{Step 3 on the worked example: solving. Left: the solved policy asks first about the fault with the highest probability per
turn. Right: the total turns each of the six orders takes over the eight past tasks, where fewer is better. Notice that the
solved order takes the fewest turns, and asking about travel last takes the most.}
\label{fig:solve}
\end{figure}

Which check should come first? One might start with the most likely fault, or with the check that takes the fewest turns.
The solved policy combines the two. It asks first about the fault with the highest probability per turn, the classical rule for
ordering tests \citep{kadane1977search}. Travel comes first at 0.75 per turn, then airplane mode at 0.5, then the network mode at 0.25. The rule is exact when there's one fault to find. With several at once it's an approximation, but here it still picks the best of the six orders (Figure~\ref{fig:solve}).

Is this order really the best? Here we can check, because there are only six ways to order three checks. A conversation ends as soon as the last of the customer's faults is gone, so a task costs the turns of every check up to that point.

Take the three customers with roaming as their only fault. Asking about travel first ends each of their conversations after one turn.
Asking about it last makes them sit through the other two checks first, four turns each. The two-turn network check
gains from going last, too: only the four customers with that fault ever pay for it. Added up over the eight past
tasks, the solved order takes 21 customer turns, and the other five orders take 22 to 30 (Figure~\ref{fig:solve},
right). These are totals. A single customer never needs more than four turns, the cost of all three checks.

An agent that never asks about travel does worse still. It can't fix six of the eight tasks, which is the mistake Qwen alone made on the real task (Section~\ref{sec:build}). Even so, it turns out that the order itself matters less than this example suggests. In the real telecom workflow, success comes from making every check, in almost any
order (Section~\ref{sec:gains}). What pays is that the solved policy is computed once, before any conversation, so the next
step can carry it out without a model.

\subsection{Step 4: run the solution as a controller around the model}\label{sec:step-run}

\begin{figure}[t]
\centering
\includegraphics{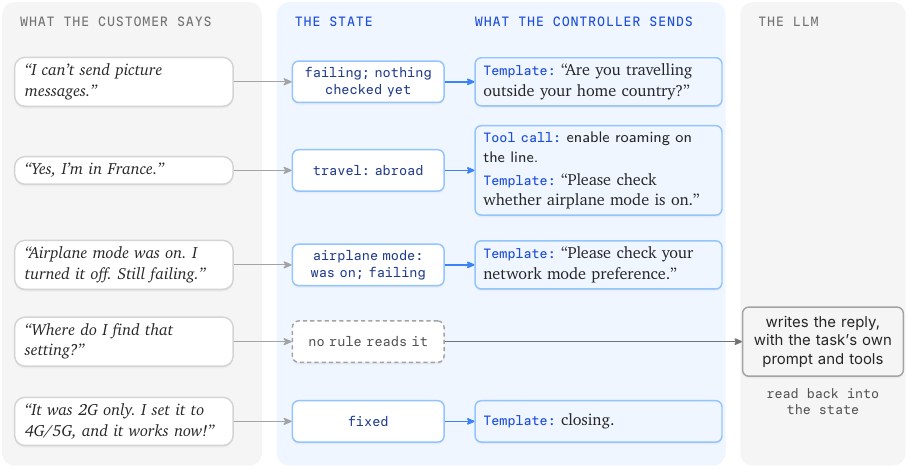}
\caption{Step 4 on the worked example: running. A new customer has all three faults. When the state fixes the move, the controller
sends it with no model call. Notice that only the off-script question goes to the LLM, and its reply is read back into the state.}
\label{fig:run}
\end{figure}

In Figure~\ref{fig:run}, we run the solved policy on a new customer who has all three faults. On every turn, the controller reads
the latest message into the state. If the state fixes what to send, the controller sends it without calling the model. We call such a turn \emph{covered}. The controller may send a templated question, an instruction, or a tool call, such as enabling roaming on the line.

When the customer says something no rule reads, such as asking where to find a setting, the turn is uncovered and goes to
the LLM. Notice that the
model itself doesn't change. It gets the task's own prompt, its tools and the whole conversation so far. The controller sends the LLM's reply and reads it back into the state, so it can take over again on the next turn.

In this conversation, the LLM writes one of the five replies, and the checks take four customer turns, as Figure~\ref{fig:solve}
counts them. That's a \blm's main saving: the other four replies cost no model call.

Because every uncovered turn goes to the LLM, the solved policy doesn't have to cover every state. It only has to be right about the turns it covers.

The controller also looks at what the LLM sends. Suppose the model tries to hand the customer to a human while checks remain and nothing is stopping a fix, as a locked SIM card would. The controller doesn't send the hand-off. It asks the model again, listing the
checks not yet made, at most three times per conversation. After that, the hand-off goes through.

Executing the moves, rather than suggesting them, matters. When our SymPy controller only suggested when the coding agent should stop, the
agent followed 6.5\% of the suggestions.

\subsection{The decision problem}\label{sec:problem}

Here is the decision problem a \blm{} solves, stated precisely. The state of Section~\ref{sec:prelim} is the \emph{observed state} $x$: discrete fields read from messages and tool results. Some things that matter can't be read directly, such as which faults are still present
on the phone. We call these the \emph{hidden state} $h$. The controller keeps a \emph{belief} $b$, a probability for each
value of $h$.

A \emph{move} $a$ costs $c(x,h,a)$. Some moves gather information, such as a lookup, a check or a second run of a failed test job. Some act, such as a fix or an order. Some end the task, such as closing or a hand-off to a person. After each move, the reader turns what comes back into an observation $o$, which occurs with probability
$P(o\mid x,h,a)$.

Choosing moves to minimize the expected total cost then becomes a partially observable Markov decision process
\citep{kaelbling1998pomdp}. When nothing that matters is hidden, it's a Markov decision process. A solution is a
\emph{solved policy} $\pi(x,b)\mapsto a$, stored as a table or computed by a small search at run time. In the worked example, for instance, $x$ holds
the checks made and the customer's reports, and $h$ is the set of faults still present. Each move costs the customer turns
it takes, and the solved policy is the order of Figure~\ref{fig:solve}.

Whoever deploys the agent chooses what the solved policy minimizes, not the model. We price every move and every outcome in one unit, and the solved policy minimizes the expected total. In telecom the unit is customer turns, so the solved
policy asks first where a fault is most likely per turn spent. Minimizing turns can't make it give up early. It closes only
once the customer reports the problem fixed. It hands the customer to a human only when the written policy says to, as for a
locked SIM card.

Other domains use other units. In continuous integration it's dollars, from runner prices and developer time. In
InventoryBench it's the benchmark's reward, the margin on sales minus the cost of holding stock. In SymPy it's billed tokens.
A different unit gives a different solved policy from the same probabilities. If every check took the same number of turns,
the order would follow the probabilities alone.

The unit can be time, too. When speed is what matters, price each move in minutes or turns, as telecom does with customer
turns. A hard time limit doesn't add up move by move, so it goes into the state as the time left, or becomes a limit on the
expected total \citep{altman1999cmdp}. Whatever the unit, the moves the controller sends cost no model call. So the saving
in model calls, and in what they cost, comes from executing the solved policy, not from what it minimizes
(Section~\ref{sec:costlaw}).

Three conditions make this work. We can list the moves. We can observe the outcomes that matter, so we can estimate their
probabilities. And a finite set of states is enough to tell apart the conversations that need different moves.

The third condition is where saturation comes in, and it depends on choosing the state well. The state has to keep what the
decision depends on, not the whole conversation. In the worked example, the checks made can come in only eight combinations. In
telecom, with 14 checks, the record of checks made keeps growing, because LLM agents try checks in new combinations.

But the state the solved policy acts on stops growing (Observation~1). In telecom, the solved policy makes the checks in a fixed order. So it doesn't need the whole record of checks made, only the first check in that order not yet made. Its state holds that check, the kind of problem, whether it's fixed yet, and anything stopping a fix. Section~\ref{sec:theory} makes the third condition precise and bounds what it costs when it holds only
approximately.

\subsection{The algorithm}\label{sec:algorithm}

Algorithm~\ref{alg:build} builds the solved policy once, and Algorithm~\ref{alg:run} runs it on every turn. Together they're
the four steps above, written for any workflow.

\begin{algorithm}[ht]
\caption{Building the solved policy, once.}\label{alg:build}
\KwIn{data $D$: the workflow's logs, or runs of the LLM agent; the moves $A$; the cost of each move}
\KwOut{a solved policy $\pi$}
$S \leftarrow \textsc{Read}(D)$ \tcp*{every conversation as a sequence of states}
$\hat P \leftarrow \textsc{Estimate}(S)$ \tcp*{probability of each outcome, by counting}
\While{the last runs added new states, or $\hat P$ is still changing}{
  $D \leftarrow D \cup \{\text{more runs of the agent}\}$\;
  $S \leftarrow \textsc{Read}(D)$; \quad $\hat P \leftarrow \textsc{Estimate}(S)$\;
}
$\pi \leftarrow \textsc{Solve}(\hat P, \text{costs})$ \tcp*{value iteration, a sort, or planning over beliefs}
\Return{$\pi$}
\end{algorithm}

\begin{figure}[ht]
\begin{minipage}[t]{0.495\textwidth}
\begin{algorithm}[H]
\caption{Running a Bellman model on a conversation.}\label{alg:run}
\KwIn{solved policy $\pi$; the agent's LLM, with the task's own prompt and tools}
\While{the conversation is not over}{
  read the latest message or tool result into $(x,b)$\;
  $a \leftarrow \pi(x,b)$\;
  \eIf{$\pi$ covers $(x,b)$ and $x$ fixes the content of $a$}{
    send $a$, a tool call or a template \tcp*{no model call}
  }{
    $m \leftarrow \textsc{LLM}(\text{prompt}, \text{history})$ \tcp*{with no LLM: a fixed request}
    \If{$m$ is a consequential move the state rules out}{
      block $m$ and ask the LLM again, up to a limit\;
    }
    send $m$\;
  }
}
\end{algorithm}
\end{minipage}\hfill
\begin{minipage}[t]{0.475\textwidth}
\begin{algorithm}[H]
\caption{The alternative: the solved policy in the prompt.}\label{alg:prompt}
\KwIn{solved policy $\pi$ written as instructions; the agent's LLM, with the task's own prompt and tools}
prompt $\leftarrow$ task prompt $+$ $\textsc{Describe}(\pi)$\;
\textcolor{red}{\footnotesize Flaw 1: the model must apply $\pi$ itself.}\;
\While{the conversation is not over}{
  $m \leftarrow \textsc{LLM}(\text{prompt}, \text{history})$\;
  \textcolor{red}{\footnotesize Flaw 2: every turn is a model call.}\;
  send $m$\;
  \textcolor{red}{\footnotesize Flaw 3: nothing makes $m$ follow $\pi$.}\;
}
\end{algorithm}
\end{minipage}
\end{figure}

Algorithm~\ref{alg:build} stops collecting when new conversations stop adding states and the estimated probabilities stop
moving. Section~\ref{sec:stopping} turns this test into a number, and Section~\ref{sec:build} shows how we chose the data for
the telecom controller. Algorithm~\ref{alg:run} reads every turn into the state, including the turns the LLM writes, so the
controller always knows where the conversation stands.

Algorithm~\ref{alg:prompt} looks similar to Algorithm~\ref{alg:run}, since both have the solved policy. The difference is where the decision happens. In Algorithm~\ref{alg:prompt} the model still reads the conversation and decides on every turn.
So every turn is still a model call, and it's up to the model whether to follow the solved policy. We test this alternative
at three levels of detail (Section~\ref{sec:arms}), and Observation~6 compares it with executing the solved policy.

\subsection{Building the telecom controller}\label{sec:build}

\begin{figure}[t]
\centering
\includegraphics{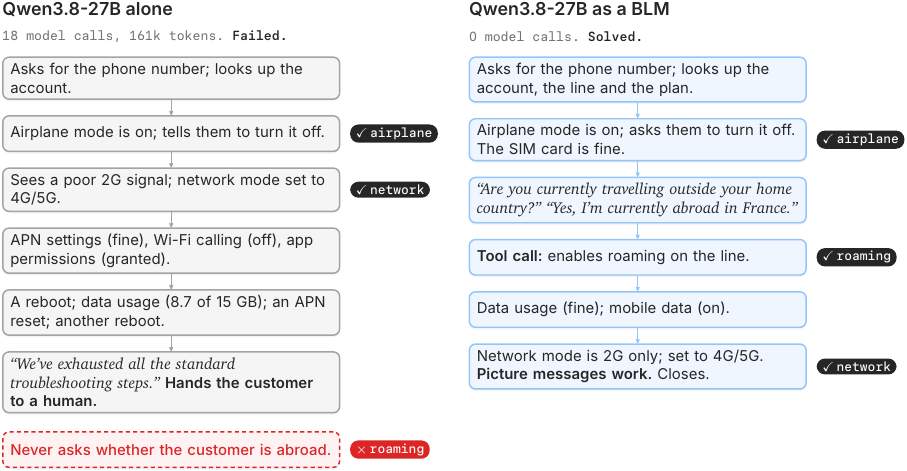}
\caption{The real task the worked example comes from, abridged from the logged transcripts. Dark tags mark each fault an agent fixed. Notice
that alone, Qwen3.8-27B never asks about travel and hands the customer to a human. In the BLM, the controller sends every move and fixes all three faults with no model call.}
\label{fig:pair}
\end{figure}

The three faults of the worked example come from a real task in our test set, which we picked as an example. Section~\ref{sec:results}
gives the rates over all tasks. Figure~\ref{fig:pair} sets two conversations on this one side by side. On the left, Qwen3.8-27B runs the benchmark's own agent. On the right, the same LLM runs as a Bellman model. The simulated
customer is the same in both.

Both agents found the airplane mode, and Qwen alone even fixed the network mode, which the Bellman model reached only later. But
Qwen alone never asked whether the customer was traveling. After checking everything else it could think of, it told the
customer it had run out of standard troubleshooting steps, and handed them to a human. The BLM asked about travel third, enabled roaming itself with one tool call, and finished the task without calling the model once.

Why third, and not first as in the worked example? In short, because the guide puts two other checks in an earlier stage. The rest of this subsection follows Algorithm~\ref{alg:build} for the controller
behind the right-hand column, the last of the three we built for telecom, and says what changes from the example.

Before building anything, we split the benchmark's 2,285 tasks once (Section~\ref{sec:protocol}). The probabilities and the costs below come only from the 1,175 training tasks. Each task is a customer with a known set of faults, as in the worked example.

We used two kinds of data. The first is the tasks themselves, which for picture messaging are 1,013 cases, each with its faults.
The second is conversations on training tasks. Claude Opus 5.5 ran on its own on 400 of them. These were 238 picture-messaging tasks, and every task of the benchmark's two other kinds of problem, a phone with no service and one with no mobile data.
Qwen3.8-27B had 1,321 conversations (Appendix~\ref{app:saturation}).

In practice, the real reader has more to read than the example's. From the customer's first message it reads the kind of problem. From each
later message, it reads whether the problem is fixed yet and what the customer reports about each setting. From the agent's own messages, it reads which check the agent asked for, and from the customer's reports and the account lookups, anything stopping a fix. The coding agent wrote the rules from the benchmark's troubleshooting guide and what the phone displays, and corrected them on training conversations and on the failures of earlier controllers.

Measured against logs the agent never sees, the reader identified the kind of problem in 1,717 of 1,721 training
conversations and whether it was fixed yet on 95.7\% of turns. It found the checks actually made with a precision of 0.90 and a recall of
0.875.

The probabilities are counts, as in the worked example. Roaming is off while the customer is abroad in 750 of the 1,013
picture-messaging tasks, so its probability is 0.74. The VPN fault occurs in none, so its probability is 0. The other two
solvers we compare (Section~\ref{sec:arms}) also count how often each combination of faults occurs. They update these probabilities after every reply by Bayes'
rule, with a fixed allowance for misreading the customer.

The turns come from Opus's conversations. A check's cost is the number of customer turns from the one where Opus first
asked about it to the one where it asked about the next. We averaged this over the 229 solved conversations among the first 233 to
finish. Asking about travel took 1.20 turns, and the SIM card took 2.86.

Solving is the same sort as in the worked example, with one change. The guide orders the checks in three stages: the line's service,
then mobile data, then picture messaging. Within each stage, the solved policy asks first about the fault with the highest
probability per turn. Checks for faults that never occur come last.

Roaming has the highest probability per turn, but comes third (Table~\ref{tab:order}). Airplane mode and the SIM card
come first, because they belong to the first stage. That's why the BLM asked about travel third. The turn counts come from
Opus's own conversations, yet Opus orders the checks differently, and asks about travel only after seven other checks.

\begin{table}[ht]
\centering\small
\caption{The solved order for picture messaging, beside Opus's. Stage: the guide's stages, in order. Fault rate:
share of training tasks with the fault. Turns: mean turns Opus spent on the check, with (1) where none were measured. Opus's
order: by when it first asks about each check, and a dash means never. Notice that roaming comes third, behind the first
stage.}
\label{tab:order}
\setlength{\tabcolsep}{2.5pt}
\begin{tabular}{@{}clcccc@{}}
\toprule
Stage & Check & Fault rate & Turns & Rate per turn & Order: solved / Opus \\
\midrule
1 & Airplane mode & 0.49 & 1.09 & 0.45 & 1 / 1 \\
1 & SIM card & 0.50 & 2.86 & 0.18 & 2 / 3 \\
\midrule
2 & Roaming (travel question) & \textbf{0.74} & 1.20 & \textbf{0.61} & \textbf{3} / \textbf{8} \\
2 & Data usage & 0.51 & (1) & 0.51 & 4 / --- \\
2 & Mobile data & 0.50 & 1.36 & 0.37 & 5 / 2 \\
2 & Network mode preference & 0.51 & 1.60 & 0.32 & 6 / 6 \\
\midrule
3 & Wi-Fi calling & 0.53 & 1.03 & 0.52 & 7 / 9 \\
3 & MMS APN settings & 0.53 & 1.10 & 0.48 & 8 / 7 \\
3 & App permissions & 0.70 & 1.84 & 0.38 & 9 / 12 \\
\midrule
-- & APN reset, reboot, line, data saver, VPN & 0 & --- & 0 & 10--14 / 10, 11, ---, 5, 4 \\
\bottomrule
\end{tabular}
\end{table}

Last, each check gets one fixed instruction, and each fault one fixed way to fix it. The account lookups and the written policy's procedures become tool calls, such as enabling roaming for a customer abroad or taking a payment before resuming a suspended
line. The controller reads their arguments from the state. These are written once, from the written policy, the guide and the tools.

In our experiments, we fixed the training data in advance rather than collecting it until nothing new appeared. Measured afterwards, the states in them had saturated (Observation~1).

\subsection{Running the controller, with and without an LLM}\label{sec:run}\label{sec:choices}

In the conversation of Figure~\ref{fig:pair}, the state fixed what to send on every turn. So the controller sent every
message, and the model was never called. Appendix~\ref{app:trace} follows that conversation turn by turn, from the
controller's side.

A turn is uncovered when the state doesn't fix what to send, as when the customer asks something off-script or answers in a
way no rule reads. In the worked example, such a turn went to the LLM, and that's one of two ways we run a BLM. Without an LLM, the
controller sends one fixed request instead. It asks, ``Could you please tell me what you see on your phone right now, so I
can help you with the next step?'' It then reads the reply like any other.

\subsection{Implementation}\label{sec:implementation}

Other domains read their states in other ways. For SymPy issues, a small fine-tuned classifier, Laya \citep{laya}, reads
which phase of its work the coding agent is in. In continuous integration and inventory control, every field of the
state comes from the system exactly. Appendix~\ref{app:domains} gives the state, the probabilities, the costs and the solvers
of each domain.

Where data are thin, we can start from an LLM's own estimates of the probabilities and let the counts correct them. The LLM's guess then counts like a fixed number of past tasks (Section~\ref{sec:prescription}).

The controllers are 300 to 1,250 lines of Python per domain. They were written by an AI coding agent, Claude Code with
Claude Opus 5.5, under our direction. It read each benchmark's instructions, tools and development logs. It wrote
the readers, templates and solvers, and worked out from the logs why runs failed. Section~\ref{sec:training} gives the point at which
building one pays for itself. At run time, a step of the telecom controller takes about 66 microseconds, timed on an earlier
version.

\section{What solving once guarantees}\label{sec:theory}

Section~\ref{sec:method} described how we build and run a \blm{}. What does it guarantee, and at what cost? Each result comes with its intuition, and proofs are in Appendix~\ref{app:proofs}.

\subsection{Optimality for the estimates}\label{sec:optimal}

Once we write down the probabilities and the costs, finding the best moves is a solved problem. Dynamic programming finds the solved policy, the one with the least expected total cost under those estimates.

Two design rules keep the problem well posed.
Costs must add over steps, so anything that doesn't add, such as a shared budget, goes into the state. And every task must be able to end, and not ending must cost something. So each state offers a move that ends the task, such as closing or a hand-off, and every other move has a positive cost, or a step limit ends any loop.

\begin{proposition}[The solved policy is optimal for the estimates]\label{prop:bellman}
If the states carry everything the decision depends on, costs add over steps, some policy ends the task with probability one
from every state, and any policy that never ends it has infinite expected cost, then value iteration over the estimated decision process converges to the optimal costs-to-go, and choosing in each state the move with the lowest cost-to-go gives a policy of least expected total cost for that process
\citep{bellman1957dp,bertsekas1991ssp}.
\end{proposition}

With a hidden state, the beliefs themselves form a decision process \citep{kaelbling1998pomdp}, which we solve
approximately, in three ways that Section~\ref{sec:arms} describes and Section~\ref{sec:analysis} compares. Proposition~\ref{prop:bellman} says nothing about how well our estimates match the world. The next results bound the two ways it can fall short: states that group too much together, and
estimates that are off.

\subsection{The cost of discretizing}\label{sec:discretize}

A decision state groups many conversations together. In telecom, conversations that made different checks, in different orders, can share a decision state. They only have to agree on the kind of problem, whether it's fixed yet, anything stopping a fix, and the first check in the solved order not yet made.
Grouping throws information away, and what we lose depends on how differently the grouped conversations would score each move. To state this, call everything about a conversation so far its \emph{full state} $s$. Write $Q^\star(s,a)$ for the least expected cost of finishing the task after taking move
$a$ in $s$.

\begin{proposition}[Discretizing costs at most $2\varepsilon$ per decision]\label{prop:abstraction}
For full states that meet the conditions of Proposition~\ref{prop:bellman}, call a grouping of them into decision states
$\varepsilon$-consistent if any two full states $s,s'$ in the same decision state satisfy
$|Q^\star(s,a)-Q^\star(s',a)|\le\varepsilon$ for every move $a$. Let $\pi$ choose, in each decision state, a move that minimizes some weighted average of $Q^\star(s,\cdot)$ over the full states it groups. Then from any start state $s_0$,
$V^\pi(s_0)\le V^\star(s_0)+2\varepsilon\,\mathbb{E}_\pi[T]$, where $V^\pi$ and $V^\star$ are the expected costs of following $\pi$ and of acting optimally, and $T$ is the number of decisions $\pi$ makes.
\end{proposition}

The proof uses the performance difference lemma \citep{kakade2002approx}. Proposition~\ref{prop:abstraction} lets us trade $\varepsilon$ against savings. Grouping more gives a smaller solved policy and larger savings, at a loss that grows linearly in
$\varepsilon$ and in the length of the task. Solving the grouped model directly is exact at $\varepsilon=0$
\citep{li2006towards} and loses an amount linear in $\varepsilon$ otherwise \citep{abel2016near}.

The proposition also suggests a rule for finding decision states from data. When the conversations grouped into one state keep needing different moves, the state groups too much: split it, as utile distinctions do \citep{mccallum1995utile}. When two states never lead to different
moves, merge them. Stop when neither happens.

\subsection{When to stop collecting data}\label{sec:stopping}

Algorithm~\ref{alg:build} adds runs until no new states appear and the estimates stop changing. We can make both parts of this test precise, the first here and the second in the next subsection.

For new states, suppose the last $n$ conversations added no new decision state, and future conversations come from the same
distribution as the collected ones. Then, with 95\% confidence, fewer than a share $3/n$ of future conversations will reach a
decision state not seen so far: this is the rule of three \citep{hanley1983zero}. Those conversations fall back to the LLM.

So if we accept that a share $p$ of conversations reach an unseen state, we can stop after $3/p$ conversations without a new one, for example 150 conversations for $p=2\%$. Observation~1 applies the rule to telecom. The rule assumes
that $n$ is fixed in advance, so when it's applied after every conversation, its 95\% level is approximate. It also says
nothing about a shift in the kind of task arriving. When the mix of tasks changes, the count starts again.

For the estimates, the next result shows that only one kind of change matters: a change large enough to flip a decision.

\subsection{How precise the estimates need to be}\label{sec:precision}

An estimate matters only if it changes a decision. If the cheapest move is the same for every probability between the
estimate and the truth, an error in the estimate costs nothing.

\begin{proposition}[Estimates matter only near decision thresholds]\label{prop:calib}
Let $Y\in\{0,1\}$ be the hidden outcome, such as whether a failure is a regression, and let each move $a$ that ends the task have cost $C(a,b)=b\,c_{a1}+(1-b)\,c_{a0}$, linear in
$b=P(Y{=}1)$ with slope $s_a=c_{a1}-c_{a0}$. Acting on a belief $\hat b$ when the true probability is $b$ costs at most
$|\hat b-b|\cdot\max_{a,a'}|s_a-s_{a'}|$ more than acting on $b$, and nothing unless a decision threshold lies between them.
\end{proposition}

For example, in Mozilla's continuous integration, a failed test job is either a regression or an intermittent failure. Marking it intermittent costs two developer-hours if it was a regression. Backing out the change costs an hour of rework if it wasn't. Escalating to a person costs fifteen minutes either way. Once rerunning the job is no longer worth its price, the
cheapest move depends only on which side of two thresholds the belief falls: 0.125 and 0.75.

Notice that an estimate of 0.3 that should have been 0.5 changes nothing. An estimate of 0.8 that should have been 0.7 backs out a change that should have been escalated
(Section~\ref{sec:calibration} shows this happening). So we should spend precision only near the thresholds. This is also why a solved policy can
start from an LLM's rough estimates, and why the second stopping test needs only the estimates near a threshold to stop changing.

\subsection{What solving once saves}\label{sec:costlaw}

Amdahl's law \citep{amdahl1967validity} says that making part of a job free can never save more than that part's share of
the work. The same holds here. The turns the controller decides become nearly free, and the turns that still go to the model
set the limit. Here we count what it costs to make the decisions, in model calls, with a decision made by the model costing one call. This isn't the cost the solved policy minimizes (Section~\ref{sec:problem}): a solved policy that minimizes customer
turns also saves model calls, because it's executed rather than re-derived.

Let reading each turn into the state cost a fraction $\rho$ of a model call, and let a share $f$ of decisions fall back to the model. If we also have the model double-check some of the
controller's decisions, to catch its mistakes, let that share be $a$. Finally, let building the solved policy cost $B$ once.

\begin{proposition}[Cost law]\label{prop:cost}
Over $N$ conversations of $T$ decisions each, solving once makes deciding cheaper by the factor
\begin{equation}
R=\frac{1}{B/NT+\rho+f+a}.\label{eq:law}
\end{equation}
It pays for itself once $NT>B/(1-\rho-f-a)$, and $R$ approaches $1/(\rho+f+a)$ as the number of conversations grows. If the solved decisions also
change the number of decisions per task to $T'$, the last three terms scale by $T'/T$.
\end{proposition}

\begin{figure}[ht]
\centering
\includegraphics[width=0.62\textwidth]{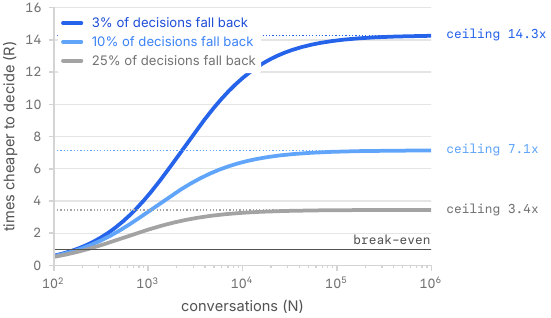}
\caption{The cost law (Proposition~\ref{prop:cost}) for an example with 120 decision states, 4 moves each, 3 decisions per conversation, $\rho=0.02$ and $a=0.02$, and building costing one model call per state and move. Each curve is a fallback share.
Notice that savings level off at $1/(\rho+f+a)$: about 14, 7 and 3 times cheaper for 3\%, 10\% and 25\% fallback.}
\label{fig:costlaw}
\end{figure}

Figure~\ref{fig:costlaw} draws the law for a small example. Savings rise quickly with traffic and then level off at a
ceiling set by the share of decisions that fall back to the model. Past that point the cost of building no longer matters,
and only the fallback share still changes the savings.

That share, which plays the role of Amdahl's serial fraction, is one we can measure before deploying the solved policy. Replay the frozen controller over an agent's existing logs and count the turns where the controller's state already fixes the content. We call this the \emph{decidable share}, and the fallback share is roughly what's left. Section~\ref{sec:results}
compares it with the calls the \blm{}s actually removed, in four domains. Section~\ref{sec:training} works out the break-even
point for the telecom results.

\section{Four domains, from nearly all decisions to nearly all content}\label{sec:setup}

Where should solving once pay off? We hypothesize that how much of an agent's work we can solve in advance depends on how much of it is choosing among
recurring moves, rather than writing new content. So we chose four domains across that range. Telecom troubleshooting is mostly decisions, inventory control and test-failure triage are only decisions, and resolving SymPy issues is mostly content.

\subsection{Domains}\label{sec:domains}

\textbf{\tautwo-bench telecom} \citep{barres2025tau2,tau2repo}. A simulated customer has a phone with one or more faults and a problem of one of three kinds: no service, no mobile data, or picture messages that won't send. The agent works on the account through tools. The customer works on the phone, so the agent has to ask for every phone check. The benchmark's own scorer judges success. The benchmark has 2,285 tasks.

\textbf{InventoryBench} \citep{inventorybench2026}. An agent orders stock each period under uncertain demand and uncertain delivery times. The benchmark has 1,320 instances built from real H\&M demand and from synthetic series, in six batches (real or
synthetic demand, with lead times of zero periods, four periods, or random). The score is the benchmark's normalized reward,
and published methods are on its leaderboard \citep{inventorybenchboard}. The controller makes no model calls here.

\textbf{Mozilla CI triage} \citep{lampel2021oranges}. Each failed test job on Mozilla's integration branch is either a
regression or an intermittent failure. The controller can retrigger the job, ask an LLM to read its error lines, mark it
intermittent, back out the change, or escalate to a person. Costs are runner list prices times minutes, and developer time:
fifteen minutes for an escalation, two hours for a missed regression, one hour for a backout that wasn't needed. From 223,699 failed
jobs over 120 days we sampled 5,000 with staff labels. Controllers are fitted on an earlier training window and tested on the 1,173 failures of a later window.

Scoring every method on every failure needs the outcomes of retriggers that never happened. So we report two scores: one on the 169 failures with all outcomes observed, and one on all failures, with the missing outcomes simulated. The score is dollars per failure.

\textbf{SymPy issue resolution} \citep{jimenez2024swebench}. A Qwen3.8-27B coding agent works on issues from the SymPy
repository. The controller decides when the agent should run its tests and when to stop, and it carries out the submission itself. It reads the agent's phase with the Laya classifier. We test on 21 issues, three runs each, scored by the
issues' hidden tests, with the agent's billed tokens as cost.

Appendix~\ref{app:domains} lists, for each domain, the state, the hidden variable, the reader, the moves, the cost unit and
the solvers.

\subsection{What we compare}\label{sec:arms}

We use two LLMs: Qwen3.8-27B \citep{qwen38}, an open-weights model that we run ourselves, and Claude Opus 5.5
\citep{opus55}. Unless we say otherwise, Qwen3.8-27B runs with its thinking mode off. Appendix~\ref{app:models} gives the details of how we ran each model.

The main comparison is always between an LLM agent and a Bellman model built on the same LLM. The LLM agent is the benchmark's own
agent, unchanged: the same prompt, written policy and tools. In the BLM, the controller sends the move on every turn the solved policy covers, and the same LLM writes every other turn. Note that the controller's moves go out as ordinary messages and tool calls, so the benchmark
scores both agents in exactly the same way.

Two variations of the \blm{} show how the work is divided. In the first there's no LLM at all: when a turn isn't covered, the
controller simply asks the customer to describe what they see (Section~\ref{sec:run}). This measures what the solved policy
does on its own. In the second, which we call the \emph{thinking fallback}, the LLM that writes the uncovered turns uses its thinking mode.

To separate knowing the solved policy from executing it, we also give the LLM the solved policy as text, as in Algorithm~\ref{alg:prompt},
at three levels of detail. Concretely, the \emph{checklist} gives the order of checks and the steps of each procedure. The checklist with templates adds the exact wording of every message the controller would send. The \emph{oracle plan} is the benchmark's own list of the actions each task requires. We also ran the checklist with thinking on.

We can solve the decision process in more than one way, and we compare three solvers, all using the same states, probabilities and costs. The simplest ignores the uncertainty about the hidden state. In telecom it's the fixed order of checks in Section~\ref{sec:build} (Table~\ref{tab:order}). In the other domains, it acts as if the most likely value of the hidden
state were the true one.

The second, QMDP \citep{littman1995qmdp}, scores each move under the current belief, but assumes the
uncertainty will be gone after one step. The third plans over beliefs, and Appendix~\ref{app:domains} gives each domain's version. In telecom, the fixed order was the configuration we registered in advance as our main one.

In the other domains we compare against the strongest baselines we could find. In InventoryBench, these are the methods on the benchmark's leaderboard and a published operations-research rule, which our implementation reproduces. We also ran Claude Opus 5.5 with the benchmark's best published strategy, in which the LLM sets each order after seeing that rule's recommendation, on 60 instances drawn at random. In CI, they're simple rules, such as
``retrigger once, then escalate'', and an LLM given the same prices and probabilities as the controller. In SymPy, the strongest simple baseline is a rule that stops the agent at its first passing test.

\subsection{Test sets and registration}\label{sec:protocol}

We split \tautwo-telecom's 2,285 tasks before building any controller: 1,175 for training, 495 for testing, and 501 that each
contain a type of fault missing from the training tasks. We used the benchmark's own 114 base tasks during development.

We froze each controller and recorded a hash of its code before drawing its test tasks at random with a public source of random numbers \citep{nistbeacon,kelsey2019beacon}. We committed the hash, the drawn tasks and the analysis plan to a public repository within a minute of the draw, before any result. We call this record a registration. When a test showed a flaw, the fix went into a new
controller, tested on newly drawn tasks, and earlier results stand. This produced five telecom test sets, which we always report
separately:
\begin{itemize}[leftmargin=*,itemsep=1pt,topsep=2pt]
\item test set 1: 80 picture-messaging and 34 mobile-data tasks, for the first controller;
\item test set 2: 114 picture-messaging tasks, for the second controller, which we repeated with Claude Opus 5.5 playing the
customer (the registered comparison uses 64 of them);
\item test set 3: 114 picture-messaging tasks, for the third controller;
\item unseen-fault set 1: 60 tasks, each with a type of fault missing from all training tasks, for the first controller;
\item unseen-fault set 2: 60 other such tasks, for the third controller, with Opus playing the customer and three attempts
per task.
\end{itemize}
Appendix~\ref{app:ledger} lists every registration and its outcome, including predictions that turned out wrong.

We couldn't test the other domains this way. In CI and SymPy, we froze the controllers before their only scoring run, but we had seen the same failures and issues while building earlier
versions. InventoryBench has no separate test data, so we chose the settings for its real-demand batches using only its
synthetic batches, and the other way round.

\subsection{Measurements}\label{sec:metrics}

In telecom, Qwen3.8-27B simulates the customer, except where we say that Opus plays the customer. In practice, the simulated customer's replies vary a little from run to run even with randomness turned off, so we compare agents task by task.

We report the share of tasks solved and, where we attempted each task three times, the share solved in all three attempts,
which \tautwo-bench calls pass$^3$ \citep{barres2025tau2}. For cost and speed we report:
\begin{itemize}[leftmargin=*,itemsep=1pt,topsep=2pt]
\item model calls: the agent's LLM calls per conversation;
\item tokens: what the agent's model reads and writes, including its thinking but not the simulated customer's;
\item cost: the dollar cost of Opus's calls at list prices;
\item waiting time: the time between each customer message and the agent's reply, added up over the conversation.
\end{itemize}
Ranges in brackets are 95\% confidence intervals from resampling tasks (in CI, groups of failures from the same code
change; in SymPy, issues). Section~\ref{sec:costlaw} defines the decidable share, and Appendix~\ref{app:decidable} describes how we measure it.

\section{The states repeat, and executing the solved policy pays}\label{sec:results}

Do the claims of Section~\ref{sec:intro} hold? Table~\ref{tab:registered} gives the registered telecom results, and eight observations follow, each tied to one of the claims. The other domains appear where they bear on a claim.

\begin{table}[t]
\centering\small
\caption{Registered \tautwo-telecom results, one row per test set, each with its own frozen controller. Each cell gives the tasks
solved, with the model calls per conversation in parentheses. In unseen-fault set 1, every task contains a type of fault missing from all training tasks. In the last row, Claude Opus 5.5 plays the customer. Notice that as BLMs, both models solve as many tasks or more, with a small fraction of the calls.}
\label{tab:registered}
\setlength{\tabcolsep}{4pt}
\begin{tabular}{@{}lcccc@{}}
\toprule
Test set (controller) & Qwen alone & Qwen + \blm & Opus alone & Opus + \blm \\
\midrule
Test set 1 (first), of 114 & 52 (17.9) & \textbf{94} (4.2) & 111 (15.6) & \textbf{114} (0.43) \\
Test set 2 (second), of 114 & 31 (18.4) & \textbf{102} (4.9) & 111 (17.0) & \textbf{113} (0.61) \\
Test set 3 (third), of 114 & 33 (20.9) & \textbf{109} (1.5) & 110 (16.9) & \textbf{112} (0.25) \\
Unseen-fault set 1 (first), of 60 & 22 (19.6) & \textbf{42} (6.4) & --- & --- \\
Test set 2, Opus customer (second), of 64 & 45 (19.9) & \textbf{63} (1.6) & 64 (17.0) & \textbf{64} (0.78) \\
\bottomrule
\end{tabular}
\end{table}

\textbf{Observation 1: Most of a bounded workflow is recurring decisions, and the state those decisions depend on
saturates.} With no model at all, the solved policy solves 90 of the 114 tasks in test set 2, against 31 for Qwen3.8-27B on its own (Table~\ref{tab:ablate}).

\begin{figure}[t]
\centering
\includegraphics[width=0.8\textwidth]{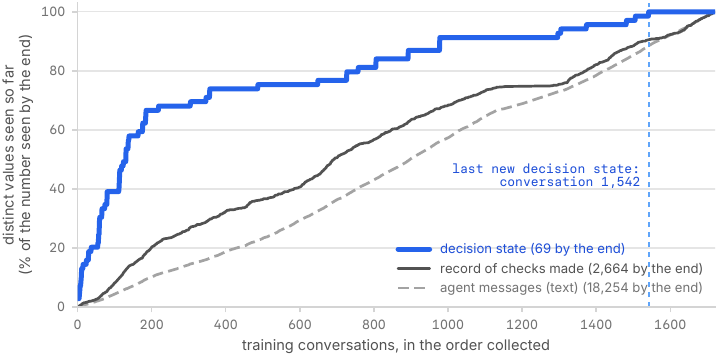}
\caption{Distinct values seen as training conversations accumulate, as a share of the number seen by the end: 1,721 telecom
training conversations that ran without a solved policy, replayed in the order they were collected through the third
controller's reader. The decision state the solved policy acts on levels off. The full record of checks made and the agent's messages keep growing to the end.}
\label{fig:saturation}
\end{figure}

Figure~\ref{fig:saturation} shows why we can decide so much in advance. We replayed the 1,721 conversations on training tasks that ran without a solved policy through the controller's reader, in the order we collected them. We counted the distinct values of three things: the decision state the solved policy acts on, the full
record of which checks a conversation had made, and the agent's messages. The decision state took 69 values, and none of the last 179 conversations added one.

The rule of three (Section~\ref{sec:stopping}) bounds the share of new conversations that
would reach an unseen decision state at 1.7\%. On the test sets, the share was 0.2\%, 2 of 965 conversations, on 5 of their 17,438 turns.

The record of checks made kept growing, to 2,664 values, with new ones in 59 of the last 100 conversations. And 518 of the 965 test conversations reached a record not seen in training: LLM agents keep trying checks in new combinations.
The agent's messages repeated least of all, with 18,254 distinct messages and new ones in 99 of the last 100 conversations.
Appendix~\ref{app:saturation} gives the make-up of the training conversations, the coverage of each test set, and further
checks of the reader.

\textbf{Observation 2: LLM agents know what they need to decide well, but they generally don't optimize.} Qwen3.8-27B's failures are
mostly hand-offs. Most of its failed conversations ended with the customer handed to a human while the solved policy still had a move to make: 49 of 62, 76 of 83 and 73 of 81 on test sets 1, 2 and 3. On test set 3, the Bellman model solved 69 of the 73 tasks Qwen lost this way.

It turns out that thinking doesn't remove the pattern: with thinking on, Qwen solves 62 of the 114 tasks in test set
2, against 102 as a \blm{} with thinking off (Table~\ref{tab:ablate}). Opus's order of checks doesn't follow the fault rates in its own conversations (Table~\ref{tab:order}).

The most direct evidence comes from a simulated coding agent, where each move costs minutes of work and we know the true probabilities, so we can score every decision exactly (Appendix~\ref{app:sim}).
There, a solved policy computed from an LLM's own estimates of the probabilities cost 29.9 minutes per task, the same as one
computed from the true probabilities. An LLM choosing each move itself cost 32.2 to 33.0 minutes.

\textbf{Observation 3: As a \blm{}, a small open model reaches the accuracy of a frontier model, including on faults it never saw.}
On test sets 1, 2 and 3, Qwen3.8-27B as a \blm{} solved 94, 102 and 109 of 114 tasks, against 52, 31 and 33 on its own
(Table~\ref{tab:registered}). Those are gains of 36.8 [27.2, 46.5], 62.3 [52.6, 71.9] and 66.7 [57.0, 75.4] points.

With the third controller, that is within one task of Claude Opus 5.5 on its own ($-$0.9 points [$-$6.1, 4.4]), at 1.5 model calls per
conversation against Opus's 16.9. With a thinking fallback, the \blm{} matched Opus at 110 tasks (0.0 [$-$4.4, 4.4]), using 4.7
thousand agent tokens per conversation against Opus's 245 thousand. The first two controllers fell short of Opus, by 14.9 and 7.9 points.

The gains held on faults the controller had never seen. In unseen-fault set 1, every task contains a type of fault missing from the training tasks, so we never estimated its probability. As a \blm{}, Qwen went from 22 to 42 of 60 tasks,
$+$33.3 points [16.7, 50.0]. The gains were largest on the tasks with the most faults, which need the most decisions (Appendix~\ref{app:faults}). They also held with a stronger customer. With Opus playing the customer on 64 tasks of test set 2, Qwen went from 45 to 63 as a \blm{}, against a registered forecast of $+$28 points ($+$28.1 [17.2, 40.6]).

\textbf{Observation 4: For a frontier model, a \blm{} removes almost all model calls, cost and waiting time, without losing accuracy.} With Claude Opus 5.5 as a \blm, the three test sets saw 114, 113 and 112 tasks solved, against 111, 111 and 110
for Opus on its own. Model calls per conversation fell from 15.6, 17.0 and 16.9 to 0.43, 0.61 and 0.25. The cost per
conversation fell from \$0.43--0.49 to \$0.012--0.024. On test sets 1 and 2, the customer's total waiting time per
conversation fell from 77.8 and 87.0 seconds to 2.5 and 3.7 seconds, with a median of 0 seconds. Part of Opus's time is the overhead of the interface we reached it through (Appendix~\ref{app:models}).

\begin{figure}[t]
\centering
\includegraphics[width=\textwidth]{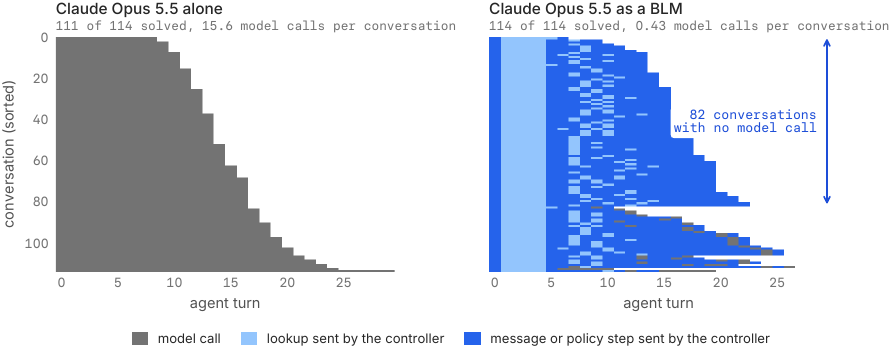}
\caption{Every agent turn on test set 1 with Claude Opus 5.5. Each row is one conversation (114, sorted by model calls) and each
cell one agent turn, colored by what produced it. Left: Opus alone calls the model on every turn. Right: notice that in the Bellman model almost every turn is an executed lookup or template, and 82 of the 114 conversations contain no model call at all. Conversations with the \blm{} have more turns, because each check is a separate instruction.}
\label{fig:barcode}
\end{figure}

Figure~\ref{fig:barcode} shows what this looks like: in 82 of the 114 conversations of test set 1, and 74 of test set 2, the \blm{} never called the LLM. Of the turns the controller executes, about a quarter are account lookups and a quarter to a third are checks in the solved order. A fifth are steps of procedures from the written policy. What didn't fall is the number of customer messages. On
test set 3, conversations with the \blm{} had 31.5 of them against 28.2 for Opus alone, because the controller asks one check per
message.

\textbf{Observation 5: \blm{}s succeed consistently across repeated attempts.} A support agent that solves a task
only some of the time is of limited use, which is why \tautwo-bench reports pass$^k$. Unseen-fault set 2 tests this. Every task contains a type of fault missing from the training tasks, Opus plays the customer, and each task is attempted three times. Qwen3.8-27B on its own succeeded in 67.2\% of attempts but solved only 40.0\% of tasks in all three
(Table~\ref{tab:holdout}).

As a \blm{}, it succeeded in 97.8\% of attempts and solved 93.3\% of tasks in all three, at 0.97 model calls per conversation. That's $+$30.6 points [22.8, 38.9] on attempts, the registered primary measure, and $+$53.3 [41.7,
66.7] on all three. With a thinking fallback, and with Opus as a \blm, every one of the 180 attempts succeeded. To save cost, we
ran Opus on its own for only 41 of its planned 180 attempts. It solved all 41, at 16.3 model calls per conversation.

\begin{table}[ht]
\centering\small
\caption{Unseen-fault set 2: 60 tasks, each with a type of fault missing from all training tasks. Claude Opus 5.5 plays the customer, and each task gets three attempts. pass$^1$: share of attempts that succeeded; pass$^3$: share of tasks solved in all three attempts. Notice that the BLMs solve nearly every task in all three attempts, at under one model call per conversation.}
\label{tab:holdout}
\begin{tabular}{@{}lcccc@{}}
\toprule
Agent & pass$^1$ & pass$^3$ & Calls & Agent tokens \\
\midrule
Qwen alone & 67.2\% & 40.0\% & 19.1 & 181k \\
Qwen + \blm{} & 97.8\% & 93.3\% & 0.97 & 11k \\
Qwen + \blm, thinking fallback & 100\% & 100\% & 0.32 & 3.2k \\
Opus alone (41 attempts) & 100\% & --- & 16.3 & 233k \\
Opus + \blm & 100\% & 100\% & 0.27 & 4.1k \\
\bottomrule
\end{tabular}
\end{table}

\textbf{Observation 6: Telling the model the solved policy improves its accuracy but not its efficiency, while executing the solved policy improves both.} On test set 2 we gave Qwen3.8-27B the solved policy as text (Table~\ref{tab:ablate}, Figure~\ref{fig:pareto}). The checklist took it from 31 to 69 tasks, the checklist with every template to 99, thinking with the checklist to 103, and the benchmark's oracle plan to 106. Every one of these still made 18 to 23 model calls per conversation, against 18.4 for Qwen on its own.

Executing the same solved policy, the \blm{} solved 102 tasks at 4.9 calls, and 112 at 0.76 calls with a thinking fallback. The
same held on the other test sets (Appendix~\ref{app:ledger}).
In other words, reading the solved policy can bring the model close to executing it in accuracy, but only at the full cost of re-deriving every turn.

Putting the decisions into the model's weights is the other way to teach them. We fine-tuned Qwen3.8-27B on the 229 solved Opus conversations from which we measured the solved order's turn counts. We evaluated it on test set 3, with a same-day rerun of the base model (Appendix~\ref{app:finetune}, not registered). The fine-tuned model solved 100 of 114 tasks at 17.6 model calls per conversation, against 109 at 1.5 for the \blm{} ($-$7.9 points [$-$15.8, $-$0.9]). The base model, rerun the same day, solved 31 at 18.9. Section~\ref{sec:training} compares the costs.

\begin{table}[ht]
\centering\small
\caption{Test set 2 (114 tasks, Qwen3.8-27B as the customer): knowing the solved policy versus executing it. Rows below the first
line are unregistered comparisons on the same tasks. Checklist: the solved check order and the procedures in the prompt;
with templates: plus every message the controller would send, verbatim. Oracle plan: the benchmark's list of the actions each
task requires. Tokens: the agent's prompt and completion tokens per conversation, in thousands, including thinking. Notice that telling the model the solved policy leaves 18 to 23 calls per conversation, while executing it leaves about 5 or fewer.}
\label{tab:ablate}
\setlength{\tabcolsep}{3pt}
\begin{tabular}{@{}lccc@{}}
\toprule
Agent & Solved & Calls & Tokens (k) \\
\midrule
Qwen alone & 31 & 18.4 & 174 \\
Qwen + \blm{} (solved order) & 102 & 4.9 & 59 \\
Qwen + \blm{} (QMDP / planning over beliefs) & 100 / 93 & 5.0 / 5.1 & 57 / 57 \\
Opus alone & 111 & 17.0 & 246 \\
Opus + \blm{} & 113 & 0.61 & 10 \\
\midrule
\multicolumn{4}{@{}l}{\emph{The policy in the prompt}}\\
Qwen + checklist & 69 & 23.1 & 232 \\
Qwen + checklist with templates & 99 & 19.4 & 206 \\
Qwen, thinking & 62 & 18.7 & 197 \\
Qwen, thinking + checklist & 103 & 19.5 & 205 \\
Qwen + oracle plan & 106 & 18.0 & 177 \\
\multicolumn{4}{@{}l}{\emph{\blm{} without the LLM, and with other orders}}\\
\blm, no model & 90 & 0 & 0 \\
Qwen + \blm, the guide's order of checks & 103 & 4.0 & 44 \\
Qwen + \blm, checks in random order & 97 & 6.5 & 77 \\
\multicolumn{4}{@{}l}{\emph{\blm{} with a thinking fallback}}\\
Qwen + \blm, thinking fallback & \textbf{112} & 0.76 & 8 \\
\bottomrule
\end{tabular}
\end{table}

\textbf{Observation 7: When the whole task is deciding, the solved policy needs little or no model. It beats every published method in inventory control, and in CI triage it costs about as much as the cheapest simple rule.} In InventoryBench, with no model calls, the \blm{} scores 0.625, against 0.538 for the best published method, and it's ahead on every batch (Table~\ref{tab:inv}). On 60
instances drawn at random, it also beats Claude Opus 5.5 running the benchmark's best published strategy, 0.598 against 0.560.

\begin{figure}[ht]
\centering
\includegraphics[width=0.7\textwidth]{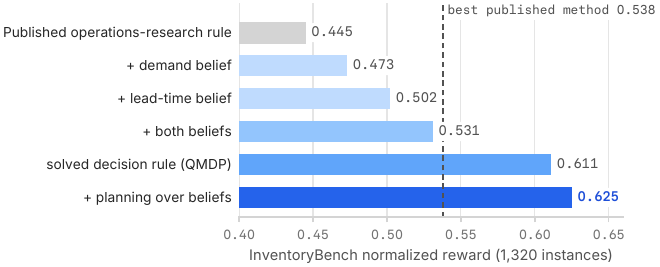}
\caption{Where InventoryBench's gain comes from. Notice that better beliefs alone bring the published rule to about the best published level, and the solved decision rule does the rest.}
\label{fig:inventory}
\end{figure}

Better beliefs and a better decision rule contribute about equally (Figure~\ref{fig:inventory}). Feeding the \blm's beliefs about demand
and lead times into the published operations-research rule lifts it from 0.445 to 0.531. The solved decision rule then takes it to 0.611, and planning over beliefs to 0.625.

In CI triage the evidence is weaker. On the 169 failures with all retrigger outcomes observed, the controller cost \$14.16 per failure. The cheapest simple rule on these failures, ``retrigger once, then escalate'', cost \$13.23, and so did Qwen3.8-27B given the same prices and probabilities. The difference from the rule isn't significant ($-$\$1.32 to $+$\$3.22 per failure). Note that these failures aren't a random sample: Mozilla had retriggered
each of them at least twice. The controller is cheaper when we simulate the missing outcomes, but simulations can favor a controller, so we don't count that as evidence (Appendix~\ref{app:ci}).

\begin{table}[ht]
\centering\small
\caption{InventoryBench normalized reward. OR rule: the benchmark's published operations-research rule, which our reimplementation reproduces.
Best published: the best score per batch on the leaderboard \citep{inventorybenchboard} (three different methods, while the best single method scores 0.538 overall). \blm{} settings were chosen on the other data split (real or synthetic) only. Notice that the \blm{} leads on every batch.}
\label{tab:inv}
\setlength{\tabcolsep}{3pt}
\begin{tabular}{@{}lcccc@{}}
\toprule
Batch & OR rule & Best published & \blm, QMDP & \blm, planning over beliefs \\
\midrule
Real, lead 0 & 0.671 & 0.772 & 0.785 & \textbf{0.786} \\
Real, lead 4 & 0.302 & 0.308 & 0.359 & \textbf{0.421} \\
Real, stochastic & 0.210 & 0.353 & 0.445 & \textbf{0.454} \\
Synthetic, lead 0 & 0.818 & 0.860 & \textbf{0.865} & \textbf{0.865} \\
Synthetic, lead 4 & 0.536 & 0.558 & 0.651 & \textbf{0.662} \\
Synthetic, stochastic & 0.107 & 0.411 & 0.519 & \textbf{0.525} \\
\midrule
Overall & 0.445 & 0.538 & 0.611 & \textbf{0.625} \\
\bottomrule
\end{tabular}
\end{table}

\textbf{Observation 8: Where the work is writing content, little can be solved in advance, and the agent's own logs show
this.} In SymPy issue resolution the hard part is writing the patch. When we replayed the controller over the coding agent's own runs, its state fixed the content of only 15 to 22\% of its turns, its decidable share. These were mostly re-runs of a test the agent had already written. In telecom, the share is 98\% of Opus's turns and 76 to 80\% of Qwen's (Appendix~\ref{app:decidable}).

The outcome followed. On 21 issues, three runs each, the \blm{} with planning over beliefs solved 19 of 63, the same as a simple rule that stops the agent at its first passing test. It read and wrote 20\% fewer tokens (127 thousand against 159 thousand per run). None of the solvers differed significantly in success from the agent on its own (14 of 63).

\begin{figure}[ht]
\centering
\includegraphics[width=0.6\textwidth]{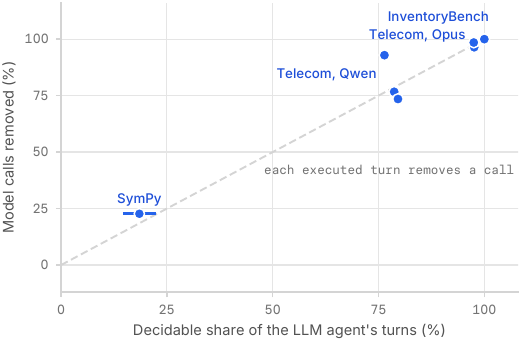}
\caption{The decidable share, measured from the LLM agent's own conversations, against the model calls \blm{} removed.
Horizontal: the share of the LLM agent's turns where a frozen controller's state fixes the content, found by replaying the
controller over those conversations (InventoryBench: 100\% by construction). Vertical: model calls per conversation removed by the \blm. Telecom: one
point per test set. The bar for SymPy spans two ways of matching the agent's turns to the controller's moves. Notice that points above the diagonal remove more calls than the share predicts, because the solved policy changes the conversation (Qwen on test set 3) or ends it early (SymPy). In CI (not
plotted), the solved policy makes every decision, though it still pays an LLM to read some logs.}
\label{fig:law}
\end{figure}

Figure~\ref{fig:law} puts the domains on one plot: the decidable share, measured from the LLM agent's own
conversations, against the model calls the \blm{}s actually removed. Notice that telecom sits near the top right, SymPy near the bottom left, and InventoryBench in the corner, where the solved policy makes every decision.

\section{Why \blm{}s work, and what still fails}\label{sec:analysis}

Where do the gains of \blm{}s come from? Which solver should we use, what still fails, and where does calibration matter? This section takes them in turn.

\paragraph{Where the gains come from.}\label{sec:gains}

Table~\ref{tab:ablate} takes a \blm{} apart on the 114 tasks of test set 2. Executing the solved policy with no LLM at all already
solves 90 tasks. Letting Qwen3.8-27B write the turns the solved policy doesn't cover brings this to 102, and letting it think on
those turns brings it to 112.

It turns out that the order of the checks matters little: the order in the benchmark's troubleshooting guide solves 103 tasks,
a random order 97, and the solved order 102. In telecom, then, the gain doesn't come from a better order. It comes from
executing: asking for every relevant check, never forgetting one, and not handing the customer to a human while one remains
(Observation~2). In InventoryBench it's different. There the decision rule itself accounts for about half of the gain
(Observation~7).

\paragraph{Which solver to use.}\label{sec:solvers}

We ran three solvers in every domain, under the same states, probabilities and costs (Table~\ref{tab:decide}). Which one did best
followed what the moves do. When moves only buy information at a price, as retriggering a job does in CI, planning over beliefs was cheapest. QMDP assumes the uncertainty will be gone after one step, so it never pays for information. When information
arrives anyway, as demand and deliveries do in InventoryBench, QMDP did well and planning over beliefs by rollout better.

When every move both tests for a fault and fixes it, as in telecom, the fixed order did as well as either. The one choice that did
badly wherever it mattered was acting as if the most likely hidden state were certain.

\begin{table}[ht]
\centering\small
\caption{The three solvers under the same states, probabilities and costs. In telecom, the simplest solver is the fixed order, which
ignores the belief. CI: dollars per failure on the 169 failures with all outcomes observed
(lower is better). In the other rows, higher is better. Notice that no solver wins everywhere: which one does best follows what the moves do.}
\label{tab:decide}
\setlength{\tabcolsep}{4pt}
\begin{tabular}{@{}lccc@{}}
\toprule
Domain (measure) & Most likely / fixed order & QMDP & Planning over beliefs \\
\midrule
CI triage (\$ per failure) & 32.79 & 16.49 & \textbf{14.16} \\
InventoryBench (reward) & 0.546 & 0.610 & \textbf{0.626} \\
SymPy (solved of 63) & 17 & 16 & 19 \\
Telecom, test set 1 (solved of 114) & \textbf{94} & 86 & 88 \\
Telecom, unseen-fault set 1 (solved of 60) & \textbf{42} & 38 & 33 \\
Telecom, test set 2 (solved of 114) & \textbf{102} & 100 & 93 \\
Telecom, test set 3 (solved of 114) & 109 & \textbf{112} & 111 \\
\bottomrule
\end{tabular}
\end{table}

\paragraph{What still fails.}\label{sec:errors}

Once we solve the decisions, the failures that remain are mostly in reading the conversation and carrying out steps. Each new telecom controller we built fixed the most common failure of the one before, and failures fell from 20 to 12 to 5 per 114 tasks.

\paragraph{Calibration matters near the thresholds.}\label{sec:calibration}

\begin{figure}[ht]
\centering
\includegraphics[width=\textwidth]{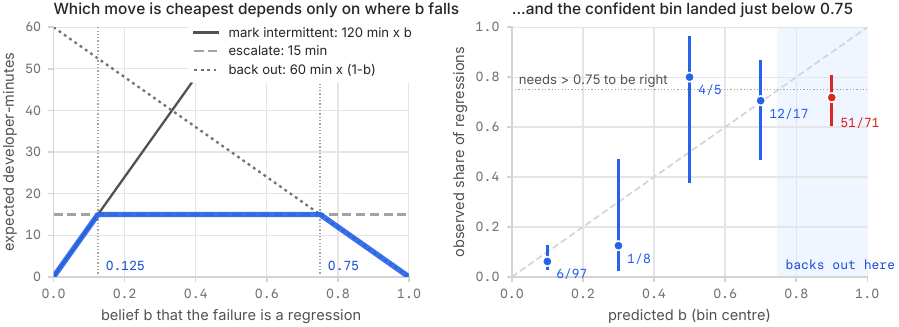}
\caption{Calibration matters only where it flips a decision. Left: expected cost of the three moves that end a CI triage, as a
function of the belief $b$ that a failure is a regression; the cheapest move changes at $b=0.125$ and $b=0.75$. Right: the
controller's beliefs before calibration, against outcomes, on 198 validation failures. Notice that its most confident bin, where it backs out changes, was right 72\% of the time (51 of 71), below the 0.75 needed for backing out to pay.}
\label{fig:threshold}
\end{figure}

Proposition~\ref{prop:calib} says that an estimate matters only near a decision threshold. CI shows the consequence
(Figure~\ref{fig:threshold}). Without calibration, the controller's most confident beliefs, where it backs out changes, were
slightly too confident and fell on the wrong side of the threshold at which backing out pays. Calibrating the weight it gives
each piece of evidence lowered its cost on the real test outcomes from \$16.18 to \$14.16 per failure (Table~\ref{tab:ci}).

\section{Building a \blm{} for a new workflow}\label{sec:prescription}

How would you build a \blm{} for a workflow of your own? The recipe is short. Use the workflow's own logs if you have them. If you don't, generate data by running the agent. From
the data, estimate the probability of each outcome and the cost of each move. Pose the decisions as a Markov decision process,
solve it, and execute the solution around the model. The eight steps below say how. Each points to the result that supports
it, and Table~\ref{tab:domains} (Appendix~\ref{app:domains}) shows how each of our domains filled them in.

\begin{enumerate}[leftmargin=*,itemsep=4pt,topsep=4pt]
\item Make sure the work is mostly deciding. List the moves the agent makes again and again, and replay a first draft of the
controller over the agent's existing logs to measure the decidable share. If most turns are spent writing new content, as in our SymPy runs, a \blm{} will save little (Observation~8).

\item Keep in the state only what the decision needs (Observation~1). Start with the fields the system reports exactly, such as tool results and the checks already made, and read the rest of the text with simple rules. Split a state when its conversations keep needing different moves, and merge states that never do (Proposition~\ref{prop:abstraction}).

\item Collect data until the decision states stop changing. If the last $n$ conversations added no new decision state, fewer than a share $3/n$ of future conversations will reach a new one. Stop once that share is small enough, since those conversations fall back to the LLM (Section~\ref{sec:stopping}, Observation~1).

\item Count outcomes in the data, and let Bayes' rule derive each state's probabilities (Section~\ref{sec:build}). Where data are thin, start from an LLM's own estimates, weighted like a fixed number of past tasks \citep{morita2008ess} (Observation~2). If an expert's conversations exist, measure costs and probabilities from them, but
don't copy the expert's choices (Table~\ref{tab:order}).

\item Price every move and outcome in one unit that whoever deploys the agent cares about, and spend precision only near the thresholds.
Take the costs from what they pay or value (Section~\ref{sec:problem}, Table~\ref{tab:domains}). Calibrate the
beliefs the controller acts on against real outcomes, near the thresholds where a decision flips (Proposition~\ref{prop:calib},
Section~\ref{sec:calibration}).

\item Choose the solver by what the moves do. Plan over beliefs when moves only buy information. Use QMDP, improved by rollout, when information arrives anyway. Order the moves by probability per unit cost when each move both tests and fixes, which is exact when there's one fault to find. And avoid acting as if the most likely hidden state were certain, as we found in Section~\ref{sec:solvers}.

\item Execute the moves the state fixes, and hand everything else to the LLM. Give the LLM the task's own prompt and tools,
read its turns back into the state, and block moves the state rules out, such as an early hand-off (Section~\ref{sec:step-run}, Observation~6).

\item Freeze the controller, test it on tasks it hasn't seen, and fix what fails. Once you've solved the decisions, the failures that remain are in reading the conversation and carrying out steps. For example, in our telecom runs, two rounds of such fixes took failures from 20 to 5 per 114 tasks (Section~\ref{sec:errors}).
\end{enumerate}

Note that none of these steps changes the model. They need the agent's logs, a reader and a solver, and the LLM keeps every turn the
state doesn't fix.

\section{Related work}\label{sec:related}

One question separates the work below from ours: who makes the agent's decisions? A person can write them, a model can generate them from text, or a model can make them on every turn, or once and then reuse them. In a Bellman model, a solver computes them from estimated probabilities and stated costs.

Separating what a conversation's state is from what to do next is an old idea in dialogue systems. Statistical dialogue
managers tracked a belief over the user's goal and planned each move with a partially observable Markov decision process \citep{williams2007pomdp,young2013pomdp}. Hybrid Code Networks combined a learned model with action templates and domain
code \citep{williams2017hcn}. We keep their split between tracking and deciding.

A second line of work puts structure around an LLM agent. Rasa's CALM has an LLM turn each user message into commands that drive business flows written by developers \citep{bocklisch2024calm}. NeMo Guardrails runs scripted or generated conversation flows \citep{rebedea2023nemo}, and StateFlow runs state machines designed by hand \citep{wu2024stateflow}. NOMOS compiles written policies into gates on tool calls \citep{nomos2026}, and PolicyGuide compiles them into workflow graphs that guide the agent \citep{policyguide2026}. In these systems the flow or the rules are written by a person or generated from text.

Closest in machinery, \citet{baheri2026controller} estimate a Markov chain over abstract states from logs and solve it once, but only to block tool calls that lead to bad outcomes. The LLM still picks every move. In a \blm{}, we solve the decisions themselves, from estimated probabilities and stated costs.

FlowBench finds that workflow knowledge in the prompt helps agents plan but leaves much room for improvement \citep{xiao2024flowbench}. A recent study finds that showing an agent its task state is unreliable. Enforcing the state helps when failures are frequent and the state can decide them, but hurts when the enforcement is wrong \citep{taskstate2026}. We find that executing a solved policy saves model calls, at similar or better accuracy, compared with describing it (Observation~6).

A third line turns an agent's behavior into code or reusable plans. ABCAgent has a frontier model write the agent's program,
which then runs without the model \citep{qi2026abcagent}. Agent just-in-time compilation has an LLM compile each task into code that is checked
against the tools' preconditions and postconditions \citep{agentjitcomp2026}. AgentJIT compiles recorded trajectories into guarded code that falls back to the LLM \citep{agentjitgithub}. Another method compiles the steps that recur in a stronger model's recorded reasoning into programs that guide a small model at every step \citep{bhat2026compiling}.

Workflow memory \citep{wang2024awm} and plan caching
\citep{zhang2025plancache} reuse workflows or plans from earlier successful runs. These systems report gains in accuracy,
speed or cost. In all of them a model makes the decisions, even when they're reused. The decisions a \blm{} reuses are ones we solve from probabilities and costs.

Routers and cascades \citep{chen2023frugalgpt,ong2024routellm,aggarwal2024automix} reduce cost by choosing which model answers
each query, sending easy queries to a cheaper one. A learned gate can also decide when a cheap agent should ask a stronger model for help. It works over a small set of abstract states, with values learned from task outcomes and a price per request \citep{sowietzki2026ask}. It controls only that one choice, and the agent's model still makes every move. These methods are complementary to \blm{}s: the turns a \blm{} hands to the LLM could be routed in the same way.

Other recent work changes how a language model is run, as we do, but so that it can read more rather than decide less. Recursive language models let the model call itself on pieces of an input too long for its context \citep{zhang2025recursive}, and context language models let it rewrite its own context \citep{shao2026context}. Both change what the model reads, and the model still makes every decision. A \blm{} changes what the model decides: the turns the solved policy covers never reach it.

Other work trains the model itself. Agent-tuning work fine-tunes smaller models on trajectories from stronger ones \citep{zeng2023agenttuning,chen2023fireact}. SCOPE-AD learns a cost-aware diagnostic strategy from Bellman targets and distills it into a Qwen model that acts as the agent \citep{scopead2026}. In other words, both make each model call better. Section~\ref{sec:training} compares this with running the solution outside the model.

Several methods help an LLM make expected-utility or information-seeking decisions \citep{liu2025dellma,hu2024uot,ding2026cta}.
Each still calls the model at every decision, to supply the probabilities, the values of outcomes, or candidate questions from which it computes the choice, or to make the choice itself.

Work on decision calibration
\citep{zhao2021decisioncal,kleinberg2023ucal,hu2024swapregret} judges predictions by the decisions they lead to, across classes of downstream tasks. Proposition~\ref{prop:calib} is the single-task case.

The theory we build on is classical. State abstraction \citep{li2006towards,abel2016near,givan2003equivalence,ferns2004metrics}
underlies Proposition~\ref{prop:abstraction}, and its rule for splitting states is close to utile distinctions \citep{mccallum1995utile}. Ordering tests by probability per unit cost is a classical result in optimal search \citep{kadane1977search}
and the basis of decision-theoretic troubleshooting \citep{heckerman1995troubleshooting}, and inventory control has strong
operations-research and learning baselines \citep{arrow1951optimal,xin2021capped,gijsbrechts2022drl}.

Work on continuous integration selects which tests to run \citep{memon2017taming,machalica2019predictive}, and work on flaky tests studies why they fail \citep{luo2014flaky,parry2021survey} and detects or ranks flaky tests and intermittent failures
\citep{kowalczyk2020apple,lampel2021oranges}. Note that we decide what to do about a failure, under explicit costs. The agent benchmarks we
use are \tautwo-bench \citep{barres2025tau2}, which extends $\tau$-bench \citep{yao2024taubench}, SymPy issues from SWE-bench
\citep{jimenez2024swebench}, and InventoryBench \citep{inventorybench2026}.

\section{Discussion}\label{sec:discussion}

Three questions remain. How does solving once compare with training the model on the same data? What does a Bellman model look like to someone who knows compilers? And what does the decision process give up?

\subsection{Solving once versus training}\label{sec:training}

The obvious alternative to a \blm{} is to train the model on the same data: fine-tune a small model on a stronger model's
conversations, or on the workflow's own logs. Training and solving have the same shape. Both pay a cost once and recover it over many conversations. They differ in what that cost pays for.

Concretely, training changes how the model decides on every turn. Solving takes the turns away from the model. A fine-tuned agent still
calls its model on every turn, so its cost per conversation and its response time stay those of the model it runs on.
Observation~6 shows both sides on the same tasks. Trained on Opus's conversations, Qwen3.8-27B learned much of what Opus does well, going from 31 to 100 of 114 tasks. But it still made 17.6 model calls per conversation, against 1.5 for the \blm.

The one-time costs differ as well. Fine-tuning Qwen3.8-27B on 207 of the 229 conversations (we held out 22 to choose the epoch) took 4.4 hours on one H200 GPU. Any change to the data, the written policy or the base model means training again. Solving the decision process is arithmetic. In telecom, for example, it amounts to ordering fourteen checks.

Building a \blm{} costs something else, the work of writing its readers, templates and the code for its procedures. Proposition~\ref{prop:cost} gives the point at which that pays off. With Opus on test set 3, the \blm{} saved \$0.48 per conversation (\$0.012 against \$0.49 for Opus alone). So a controller that costs $B$ to build pays for itself after $B/0.48$
conversations: about 210 conversations for a build that costs \$100, and 2,100 for one that costs \$1,000.

Some properties of a solved policy don't depend on how accurate the agent is, and training doesn't provide them:
\begin{itemize}[leftmargin=*,itemsep=2pt,topsep=2pt]
\item When we solve the decision process exactly, the solved policy is optimal for the estimates, and Propositions~\ref{prop:bellman} and~\ref{prop:abstraction} bound what discretizing costs. A trained model comes with no such statement.
\item We can read the solved policy decision by decision (Table~\ref{tab:order}) and test it against a specification. Trained weights can't.
\item Probabilities and costs are separate inputs, so we handle a change in costs by solving again, with no new data. A trained model needs training again.
\item We can measure how much the solved policy will save from existing logs, before deploying it (Section~\ref{sec:costlaw}).
\end{itemize}
The two aren't exclusive. We can use a fine-tuned model as a \blm's fallback, writing the turns the solved policy doesn't cover.

\paragraph{An analogy with compilers.}\label{sec:compiler}

Readers who know compilers will recognize the arrangement. An LLM agent is an interpreter: it holds its decisions implicitly,
in a model and a prompt, and carries it out by re-deriving each decision as it goes. Discretize, Then Decide compiles the decisions that recur. Like a tracing just-in-time compiler \citep{gal2009tracing}, it compiles only the paths that come back again and again, and guards each one. When a guard fails, it falls back to the interpreter, a step known as deoptimization
\citep{holzle1992deopt}. The analogy also says where
the savings stop: what never comes back stays interpreted, which is the limit of Section~\ref{sec:costlaw}.

\subsection{What the decision process gives up, and why the LLM stays}\label{sec:shape}

Any way of choosing among recurring moves that separates what's believed from what's wanted, and picks the move with the best expected outcome, is a decision process. Given the estimates, solving it exactly loses nothing. What can be lost is what that form assumes. The states must summarize the part of the history that matters.

Expected cost ignores limits on other costs and on risk, which decision processes with constraints handle by limiting expected costs \citep{altman1999cmdp}. Rules that must always hold can be enforced by removing the moves that break them. A solved
policy is also fixed, so people who deal with the agent often could learn how it decides, and exploit that. And a frozen
solved policy stops learning, which posterior sampling
addresses \citep{osband2013psrl}.

One limit we can't handle this way. A solver chooses only among moves that someone listed, and only the LLM can invent a new
one. When the LLM keeps handling some state in a new way that works, that way becomes a candidate move for the next version of the solved policy. This is why the LLM stays in the loop: it handles what's new, and shows what we should solve next.

\section{Limitations and future work}\label{sec:limits}

Where might these results not carry over? Our strongest evidence comes from one benchmark generator. \tautwo-bench's telecom tasks combine 19 fault types, and LLMs simulate its customers. Test sets drawn at random after we froze each controller protect against choosing tasks that favor the controller, not against a different generator. And the telecom results are one attempt per task except on unseen-fault
set 2.

The other domains meet a lower standard (Section~\ref{sec:protocol} and Observation~7). In InventoryBench, we developed the controller's demand forecasts and its rollout while looking at results on the whole benchmark. On two retail benchmarks, where the agent alone already solved most tasks, our controllers gained nothing (Appendix~\ref{app:ledger}).

A coding agent wrote the controllers under our direction (Section~\ref{sec:implementation}), and we haven't tested
a controller built from the recipe of Section~\ref{sec:prescription} alone. We applied the stopping rule of Section~\ref{sec:stopping} after collecting the training data, rather than using it to decide how many to collect.

Four directions follow. The first is how to discretize optimally. Choosing the states is an optimization problem of its
own: fewer facts make the states repeat sooner but can merge conversations that need different moves, and more facts do
the opposite. Proposition~\ref{prop:abstraction} bounds what a given choice of states costs, and the split-and-merge rule
of Section~\ref{sec:prescription} improves one, but neither finds the best.

The second is to build controllers from the recipe alone, by a coding agent with no person in the loop, which would show what a controller costs to build. The third is to add finished conversations back into the counts, so that a deployed solved policy keeps learning. Last, a solved policy is a finite object, so we can test it against a specification. For example, we can count in what share of states it makes an important move without a person, as is done for access-control rules \citep{eiers2022permissiveness}. We pursue this separately.

\section{Conclusion}\label{sec:conclusion}

An agent in a bounded workflow keeps meeting the same states, and an LLM agent re-derives each of them whenever it meets them. Discretize, Then Decide treats those decisions as an optimization problem, solves it once from the workflow's own
data, and leaves the model only what the states don't cover.

In telecom support this brought a small open model within one task of a frontier model at 93\% fewer
model calls, and let the frontier model drop almost all of its calls. In inventory control, a solved policy that calls no model beat every published method. Where the work is writing rather than deciding, the agent's own logs show that there's little to
gain. Much of what agents do in bounded workflows is deciding, and deciding can be solved once.

\section*{Use of AI assistance}
The authors used AI assistance in this work. It helped with the experiments, including writing the controllers and the
analysis code under the authors' direction, and with drafting the manuscript. All ideas and arguments in this paper are the
authors' own. All content has been checked and verified by the authors to be correct and original, and the authors take full
responsibility for it.

\bibliographystyle{plainnat}
\bibliography{refs}

\appendix
\section{Registrations and their outcomes}\label{app:ledger}\label{app:reg}

This appendix lists every registration we made and its outcome, including targets that were missed and forecasts that were
wrong. Seven registrations are in a public GitHub repository, \url{https://github.com/adarsh-vatsa/dtd-preregistration}. Each
commit holds the analysis plan, any forecasts, and the frozen files or their SHA-256 hashes, and, where tasks were drawn, the public random number used for the draw and the drawn tasks. GitHub's push records confirm the commit times given below. The other registrations were recorded locally before their runs.

In the repository, the three telecom controllers are named
v5c, v5e and v5f, test sets 1 to 3 are draws 1 to 3, and the unseen-fault sets are holdouts. Intervals are 95\% intervals from
resampling tasks, in points of success rate. Cost forecast errors are (forecast $-$ actual)/actual.

The 495 telecom test tasks were held in reserve and drawn in turn. A first set of 114, which we call test set 0, served an earlier controller that the body doesn't report (below). It took all four of the reserve's no-service tasks, 36 mobile-data
tasks and 74 picture-messaging tasks. Test set 1 took the 34 mobile-data tasks that remained and 80 picture-messaging tasks,
which left only picture-messaging tasks for test sets 2 and 3. Test set 3 was drawn from the 153 tasks left after test sets 0 to
2. Unseen-fault set 1 was drawn from all 501 unseen-fault tasks, and unseen-fault set 2 from the 441 left after it.

\subsection*{The telecom results}
\begin{itemize}[leftmargin=*,itemsep=3pt,topsep=2pt]
\item \textbf{Test set 1 and unseen-fault set 1, first controller (public; commit e969d50).} On both sets we compared Qwen3.8-27B alone against Qwen3.8-27B as a \blm{} under each of the three solvers. On test set 1 we also compared Claude Opus 5.5 alone against Opus as a \blm{} (fixed order). Primary comparison, Qwen as a \blm{} against Qwen alone on test set 1: $+$36.8 points [27.2,
46.5], met. Stated target, Opus alone's level: not met, $-$14.9 [$-$22.8, $-$7.0]. On unseen-fault set 1, $+$33.3 [16.7,
50.0]. QMDP and planning over beliefs against Qwen alone: $+$29.8 and $+$31.6 on test set 1, $+$26.7 and $+$18.3 on unseen-fault
set 1.

The random number used for the draw was published 47\,s before the controller's files were hashed, and the runs started
28\,s before the registration was committed. Conversations that ended in an infrastructure error, with no outcome, were rerun
once with the same code (84 of the 114 for Opus alone and 17 for Opus as a \blm). No rerun rule had been registered.

\item \textbf{Test set 2, second controller (public; b0163d9).} The same comparisons as on test set 1. Primary comparison, Qwen
as a \blm{} within 5 points of Opus alone: not met, $-$7.9 [$-$14.9, $-$1.8]. Against Qwen alone: $+$62.3 with the fixed order,
$+$60.5 with QMDP and $+$54.4 with planning over beliefs. Opus as a \blm{} against Opus alone: $+$1.8 [$-$1.8, 5.3].

The registration gives as its reason that one flaw of the first controller caused 16 of its 19 failures. That count was taken before the run had finished, and the final count is 11 of 20. The first random number we took was 3\,s older than the freeze, so before any test ran the tasks were drawn again from one published two minutes later, as the registration records. As on test set 1, conversations without an outcome were rerun (87 of Opus alone's 114), and three Qwen runs were moved to another machine during the run.

\item \textbf{Test set 2 with Claude Opus 5.5 as the customer, second controller (public; 99f62aa).} Registered after an
unregistered trial on the first 50 tasks of test set 2, which supplied the forecast. The other 64 tasks are the confirmatory
set: their outcomes with the Qwen customer were known, and none with the Opus customer. Primary comparison met: Qwen as a \blm{}
solved 63 of 64 against 45 alone, $+$28.1 [17.2, 40.6], against a forecast of $+$28 (forecast success about 95\% and 72\%,
realized 98\% and 70\%), at 1.6 calls per conversation against 19.9 (forecast below 2 and above 15).

On all 114 tasks, Opus alone and Opus as a \blm{} each solved 114, at 17.1 and 0.68 calls per conversation. The forecast, within 3 points of Opus alone with fewer than 1.5 calls, was met. Qwen with thinking solved 103, and Qwen with thinking and the checklist solved 114, at 17.3 calls. Some comparisons were started early or resumed at higher concurrency to use free machines, and two conversations cut off by an outage were rerun once. Code and settings didn't change.

\item \textbf{Test set 3, third controller (public; 51db041).} Built after studying the second controller's failures on test
sets 1 and 2 and on training conversations. It was frozen at 19:58:29 UTC on 9 October 2026, and the tasks were drawn with the first published random number at least 60\,s later (20:00:00). The registration was committed at 20:00:40, and the runs started at 20:00:51. We compared Qwen alone, Qwen as a \blm{} under each solver and with a thinking fallback, Qwen with thinking and the checklist, and Opus alone and as a \blm. Primary comparison met: Qwen as a \blm{} solved 109 of 114 against 33 alone, $+$66.7 [57.0, 75.4].

Targets, all met. Qwen as a \blm{} came within 5 points of Opus alone, which solved 110 ($-$0.9 [$-$6.1, 4.4]). The thinking fallback came within 2 points (110 solved, 0.0 [$-$4.4, 4.4]). Qwen and Opus as \blm{}s made fewer than 2 and 1 calls per conversation (1.47 and 0.25).
QMDP solved 112, planning over beliefs 111, Opus as a \blm{} 112, and Qwen with thinking and the checklist 105, at 19.1 calls (no
forecast). The forecasts, each followed by its outcome, were Qwen alone 25--35\% (29\%), Qwen as a \blm{} about 106 (109), the thinking fallback about 110 (110), Opus alone about 111 (110) and Opus as a \blm{} about 113 (112). QMDP and planning over beliefs were forecast within 5 points of the fixed order (met: $+$2.6 and $+$1.8).

Customer messages per conversation: 30.0 (Qwen alone), 33.7 (fixed order), 30.9 (QMDP),
31.7 (planning over beliefs), 31.4 (thinking fallback), 28.2 (Opus alone) and 31.5 (Opus as a \blm). Failures: 5 for Qwen as a \blm, 4 for the thinking fallback, 4 for Opus alone and 2 for Opus as a \blm. No conversation
ended in an infrastructure error.

\item \textbf{Unseen-fault set 2, third controller, with Opus as the customer and three attempts per task (public; b66bcdb).}
Registered after the primary test set 3 results, with the third controller unchanged. The files were hashed at 22:56:39 UTC on 9 October, and the 60 tasks were drawn with the first published random number at least 60\,s later (22:58:00). The registration was committed at 22:58:39, and the runs started at 22:58:43. We compared Qwen alone, Qwen as a \blm{} with thinking off and with a thinking fallback, Opus alone, and Opus as a \blm. Primary comparison met: on the share of attempts solved, Qwen as a \blm{} 97.8\% against 67.2\% alone,
$+$30.6 [22.8, 38.9].

Targets. The pass$^3$ of Qwen as a \blm{} was at least 25 points above Qwen alone's, as targeted (93.3\% against 40.0\%, $+$53.3 [41.7, 66.7]). Qwen and Opus as \blm{}s made fewer than 2.5 and 1 calls per conversation, as targeted (0.97 and 0.27). Two targets, Qwen as a \blm{} within 5 points of Opus alone and the thinking fallback within 3 points, weren't evaluated as registered, because Opus alone was stopped after 41 of its 180 attempts to limit cost. It solved all 41, and further attempts could only have lowered its rate,
so stopping early couldn't have favored these two targets. The thinking fallback and Opus as a \blm{} solved all 180 attempts,
at 0.32 and 0.27 calls per conversation.

The forecasts, each followed by its outcome, were Qwen alone 55--70\% of attempts (67.2\%), Qwen as a \blm{} about 95\% (97.8\%), the thinking fallback about 96\% (100\%), Opus alone about 97\% (41 of 41 attempts run) and Opus as a \blm{} about 98\% (100\%). We also forecast pass$^3$ 5 to 15 points below pass$^1$ for Qwen alone (missed: 27 points below), and 0 to 6 points below as a \blm{} (met: 4.5 for Qwen as a \blm, 0 for the others). Customer messages per conversation: 28.4 (Qwen alone), 31.4 (Qwen as a \blm)
and 29.8 (thinking fallback). The Opus service, which played the customer, was down for about a day during the runs. As registered, attempts that ended in an infrastructure error were rerun once with the same code: 128, 112, 107 and 131 of 180 for Qwen alone, Qwen as a \blm, the thinking fallback and Opus as a \blm. Opus alone was stopped after 12 of its 151.
\end{itemize}

While four comparisons on test set 2 were running, before any had finished, we fixed in writing (9 October 2026, 16:15 UTC) how the paper would be framed. The four were Qwen with the checklist, with thinking, with thinking and the checklist, and with the oracle plan. If the best of them solved fewer than 67 of the 114 tasks, halfway between Qwen alone (31) and Qwen as a \blm{} (102), the paper would lead with the difference between knowing a solved policy and executing it. Otherwise it would lead with cost and report the telecom accuracy gain as a decomposition. The oracle plan solved
106, so the paper leads with cost.

\subsection*{The other domains}
\begin{itemize}[leftmargin=*,itemsep=3pt,topsep=2pt]
\item \textbf{InventoryBench, Claude Opus 5.5 baseline (local, no forecast).} The plan, the harness and the primary comparison were hashed before any run. The comparison is the \blm{} as reported against Opus running the benchmark's best published strategy, in which the LLM sets each order after seeing the operations-research rule's recommendation, paired over 60 instances. Result: 0.598 against 0.560, a difference of $+$0.038 [0.016, 0.062]. Against Opus running the benchmark's LLM-only agent, it was 0.598 against 0.559.

NIST's public random numbers had stopped for five hours, so after a dated amendment the 60 instances were drawn from the first number published by drand, the League of Entropy's public random numbers. An earlier draw from an old NIST number was discarded unused. The benchmark's agent scripts ran unchanged, except that their API
client pointed to our endpoints and SciPy's normal quantile came from an equivalent function in Python's standard library.
Fifteen Opus runs were run again in full. Seven were cut short by a network outage, and eight never finished because the benchmark's own log extractor failed, after which we added a guard (no decision on a log that can't be parsed). Before any comparison finished, we changed the scoring script to read the published results' file paths and to compute the pre-specified paired interval. The analysis didn't change.

\item \textbf{SymPy (local, no forecast).} A first test was withdrawn because of a bug: the decision table, fitted again, put every state under one merged issue type, while the controller looked states up under each issue's own type. After the fix, two versions
were frozen before any run: the fixed controller, and a variant in which every solver defers to the agent when its two best moves
are close. Both were tested on the withdrawn test's 21 issues, three runs each.

Fixed controller: planning over beliefs solved 19
of 63, the most-likely-state solver 17 and QMDP 16, against 14 for the agent alone and 19 for the rule that stops at the first passing test. The most-likely-state and planning controllers mostly tell the agent to run a test before its first edit,
QMDP mostly defers to it, and all three end most runs by executing the submission. Variant: most-likely-state 16, QMDP 18, planning over beliefs 15. No controller's success differs significantly from the agent alone's (every interval
includes or touches zero), and every controller's token saving is significant. One fact the controller reads is whether the
agent's script that reproduces the issue passes. On pilot runs, a pass was wrong a third of the time (15 of 45).

\item \textbf{CI triage (two public registrations; 607ba1e and its amendment 6cc735b).} The controller we report and the
recalibrated controller (Table~\ref{tab:ci}) were each hashed before their single scoring run on the test window, and registered
afterwards, in the amendment. The hashed file of the first was later edited in place into the second, so its hash record and its
results remain, but the file doesn't. The two registrations also forecast the cost per failure on failures from 7 October 2026 on, which hadn't been collected or viewed when each was registered. The first, for the simpler controller, forecast \$8.15 against \$13.18 for the best rule, both computed on the failures its model was fitted on. The second forecast \$9.51 for the recalibrated controller (range \$7.14--11.89), its score on the test window. These forecasts are scored after 23 October 2026 and aren't part of this paper.
\end{itemize}

\subsection*{Registrations not reported in the body}
\begin{itemize}[leftmargin=*,itemsep=3pt,topsep=2pt]
\item \textbf{An earlier telecom controller that only blocked hand-offs (local).} Tested on the benchmark's 114 base tasks, two
attempts each, and on test set 0. Against a controller that only executed account lookups: $+$13.6 points [7.0, 20.2], against a
forecast of $+$10.5, which met the registration's second criterion, that the forecast fall within the realized interval. The
forecast put none of the gain on mobile-data tasks, where the realized gain was 15.3 points. Against Qwen alone: $+$11.8
[3.9, 19.3] on the base tasks and $+$20.2 [10.5, 29.8] on test set 0. Against a simpler rule taken from the written policy,
which blocks every hand-off unless the reader sees a reason in the written policy for it, such as a locked SIM card:
$-$3.1 [$-$9.6, 3.5], so no better.

The token forecast missed: $+$32,103 billed
tokens per conversation, agent and simulated customer together, against $+$18,375. Test set 0 was drawn with a random number published 78\,s before the controller's file was frozen, and a later, unregistered controller was also evaluated on it.

\item \textbf{SymPy, an earlier controller that suggested or executed stops (local; 20 issues, three runs each).} When the
controller only suggested stopping, the agent followed 6.5\% of the suggestions, and the saving fell short of its forecast
(9.1\% against 42.4\%). A version that executed the stop, added by a written addendum before its runs, cut billed tokens by
36.5\%, at 40.0\% against 43.3\% success (interval $-$18.3 to $+$10.0). Of the saving, 54.6\% came from 12 runs that gave up at the first step.

\item \textbf{$\tau$-bench retail, a different benchmark from \tautwo-bench retail (local; 115 tasks, two attempts each).} No controller differed
from the agent alone (61.7\% against 60.0\%, interval $-$5.7 to $+$9.1). The success forecast missed by 9.4 points and the cost
forecast by $-$16.6\%. The forecast for the agent alone was written after it ran.

\item \textbf{\tautwo-bench retail (local, no forecast).} It was frozen at 13:24:48 UTC on 8 October 2026, and the runs began 7\,s later, on
the 40 unused test tasks, two attempts each. The agent alone solved 66 of 80; as \blm{}s, the most-likely-state solver solved 61,
QMDP 66 and planning over beliefs 66. This controller left the matching of items, variants and payment methods to the LLM, and
the agent alone already solved most tasks, which left little room for a gain.

\item \textbf{Checks of two fitted models before evaluation (local).} A simulated SymPy world predicted the agent's success and
cost on held-out issues (0.385 against 0.400; cost $+$0.2\%), but failed its calibration and Markov-structure criteria. The final
$\tau$-bench retail decision table predicted the agent's success and cost (0.636 against 0.600; cost $-$1.2\%), but failed 3 of
its 5 criteria before any controller run.
\end{itemize}

\section{Proofs}\label{app:proofs}
\paragraph{Proposition~\ref{prop:abstraction}.} Write $A^\star(s,a)=Q^\star(s,a)-V^\star(s)\ge0$, with $V^\star=0$ once the task
has ended. If $\mathbb{E}_\pi[T]$ is infinite the bound is vacuous, so let $\mathbb{E}_\pi[T]<\infty$; then telescoping
$\sum_{t<T}[c(s_t,a_t)+V^\star(s_{t+1})-V^\star(s_t)]$ and taking expectations gives the performance difference lemma
\citep{kakade2002approx} in its undiscounted form, $V^\pi(s_0)-V^\star(s_0)=\mathbb{E}_\pi\big[\sum_{t<T}A^\star(s_t,a_t)\big]$.
Let $s$ lie in decision state $z$, let $\bar Q(z,a)=\sum_{s'\in z}w(s')Q^\star(s',a)$ for weights $w$ on $z$, and let
$\hat a=\arg\min_a\bar Q(z,a)$. By $\varepsilon$-consistency $|Q^\star(s,a)-\bar Q(z,a)|\le\varepsilon$ for every $a$, so
$Q^\star(s,\hat a)\le\bar Q(z,\hat a)+\varepsilon\le\bar Q(z,a^\star(s))+\varepsilon\le Q^\star(s,a^\star(s))+2\varepsilon$, that
is, $A^\star(s,\hat a)\le2\varepsilon$ at every state. Summing over the $T$ decisions gives the bound.\hfill$\square$

\paragraph{Proposition~\ref{prop:calib}.} With $\hat a=a^\star(\hat b)$ and $a=a^\star(b)$,
$C(\hat a,b)-C(a,b)=[C(\hat a,b)-C(\hat a,\hat b)]+[C(\hat a,\hat b)-C(a,\hat b)]+[C(a,\hat b)-C(a,b)]\le(b-\hat b)s_{\hat a}+0+(\hat
b-b)s_a\le|\hat b-b|\max_{a,a'}|s_a-s_{a'}|$. If no threshold separates the beliefs, $\hat a=a$.

\section{Models and settings}\label{app:models}
Qwen3.8-27B is served with vLLM \citep{kwon2023vllm} at temperature 0 with thinking off, unless stated. With thinking on, we use the model's recommended sampling (temperature 0.6, top-$p$ 0.95). In SymPy, the coding agent samples at temperature 0.6 with
thinking off, so its three runs on each issue differ. 

Claude Opus 5.5 is reached through Claude Code's command-line interface, behind an adapter that makes it behave like an ordinary chat model. The benchmark's prompt replaces the interface's system prompt, the interface's own tools are turned off, and tool calls are written as text. The interface's default reasoning is kept: 61\% of the Opus-alone agent calls on test sets 1--3 used thinking tokens. The interface adds a
fixed overhead to every Opus call: a minimal call through it, with a one-word reply, took 2.6\,s (measured with Claude Sonnet
5.5). So a native interface would cut the waiting times we report for Opus alone, but not remove them. In telecom, the simulated
customer is Qwen3.8-27B at temperature 0, except in the replications where Claude Opus 5.5 plays the customer.

\section{The fine-tuning baseline}\label{app:finetune}
We fine-tuned Qwen3.8-27B on the same conversations from which the solved order's turn counts were measured: the 229 solved
Claude Opus 5.5 conversations on training tasks. Each conversation was rebuilt exactly as the agent saw it: the benchmark's own system prompt and tool definitions, the customer's messages, and the agent's messages, tool calls and tool results. The customer's own actions on the phone aren't visible to the agent, so we left them out.

We held out 22 conversations
to choose the epoch and trained on 207. Training used LoRA \citep{hu2022lora} with rank 16 on the attention, linear-attention and MLP projections, which gives 108.8 million trainable parameters (0.40\% of the model). We used AdamW at a learning rate of $10^{-4}$ with cosine decay, eight conversations per update and three epochs, with the loss on the agent's turns only and thinking off, as at evaluation. It ran on one H200 GPU. The fine-tuned model was evaluated on test set 3 with the same harness and settings as the
registered Qwen-alone arm, with a same-day rerun of the base model as a control. This comparison wasn't registered.

Training took 4 hours 25 minutes on the GPU, including validation and merging the adapter into the model. The third epoch had
the lowest validation loss (0.342) and is the one evaluated. On test set 3 it solved 100 of 114 tasks, at 17.6 model calls
and 163 thousand agent tokens per conversation. The base model, rerun the same day, solved 31 at 18.9 calls and 184 thousand
tokens (the registered run solved 33).

\section{How each domain fills in the decision problem}\label{app:domains}
For each domain, Table~\ref{tab:domains} lists what the state holds, what's hidden and where its prior comes from, and how text or data is read into the state. It also lists the moves, the cost unit, and how each of the three solvers is implemented.

\begin{table}[ht]
\centering\scriptsize
\caption{How each domain fills in the decision process. ``Info.'' = information moves. Telecom and SymPy run live with
an LLM; CI and InventoryBench are evaluated offline.}
\label{tab:domains}
\setlength{\tabcolsep}{3pt}
\resizebox{\textwidth}{!}{%
\begin{tabular}{@{}>{\raggedright\arraybackslash}p{1.3cm} >{\raggedright\arraybackslash}p{2.4cm} >{\raggedright\arraybackslash}p{2.5cm} >{\raggedright\arraybackslash}p{1.9cm} >{\raggedright\arraybackslash}p{2.55cm} >{\raggedright\arraybackslash}p{1.2cm} >{\raggedright\arraybackslash}p{3.8cm}@{}}
\toprule
Domain & Observed state $x$ & Hidden $h$ (prior from) & Reader & Moves & Cost unit & MDP / QMDP / POMDP as implemented \\
\midrule
\tautwo-telecom & kind of problem, fixed yet, checks left, what stops a fix; line, bill, usage, travel status & remaining fault set (fault-set frequencies per kind of problem, train split) & regular expressions; tool results & info.: lookups, travel question, re-asks; acting: check
templates, payment, resume, refuel, roaming; terminal: hand-off, close & customer turns per check & fixed order by
$P(\text{fault})/$turns / check most likely to address a remaining fault / two-move lookahead \\
CI triage & platform, duration, suite, test label history, retrigger counts & regression vs intermittent (logistic prior,
train window) & retrigger outcome (beta-binomial); LLM log verdict & info.: retrigger, LLM read; terminal: mark intermittent,
back out, escalate & dollars & act on $b\ge 0.5$ / best final move under $b$ / two-move search over beliefs \\
Inventory\-Bench & on hand, outstanding orders and ages, demand history & lead-time distribution, arrival times, demand model
& exact arrivals and demand & acting: order quantity (information arrives regardless) & reward $p\cdot$sold $-\,k\cdot$stock &
most likely arrivals and demand model / newsvendor quantity at arrival over the belief / rollout over sampled futures \\
SymPy & last test result, attempts, scope, step & agent's phase before its first edit & Laya; frozen reproduction test &
info.: search, read, run, defer; acting: edit; terminal: submit (executed) & billed tokens & table of move values for the most likely phase /
move values weighted by the belief / defer when the expected value of perfect information exceeds 1,300 tokens, the cost of one agent step \\
\bottomrule
\end{tabular}}
\end{table}

\section{The Bellman model's conversation, turn by turn}\label{app:trace}

Table~\ref{tab:trace} follows the \blm's conversation of Figure~\ref{fig:pair} from the controller's side. At every turn, the
controller reads what came in, updates the state, and sends the move the solved policy picks if the state fixes its content.
Here it always did. The customer's first message set the kind of problem, and a template asked for the phone number. The
controller then looked up the account, the line and the plan, reading each tool call's arguments from the previous result. It
sent the checks in the solved order, one template each, and read every reply. A refusal to do two things at once got a re-ask. ``Yes, I'm currently abroad in France'' set off the roaming procedure, an executed tool call followed by an instruction. And ``It's working now!'' ended the conversation with a closing template. The model was never called.

\begin{table}[ht]
\centering\small
\caption{The Bellman model's conversation of Figure~\ref{fig:pair}, from the controller's side (abridged from its log). State: the kind of problem, whether it's fixed yet, and the first check in the solved order not yet made; nothing stopping a fix was found. Every message on the right
was sent by the controller, and none came from the model.}
\label{tab:trace}
\begin{tabularx}{\textwidth}{@{}>{\raggedright\arraybackslash}X >{\raggedright\arraybackslash}p{3.6cm} >{\raggedright\arraybackslash}X@{}}
\toprule
What came in & State & What the controller sent \\
\midrule
``I'm having trouble sending MMS messages\ldots'' & picture messages; unknown; airplane mode & Template: asks for the phone
number. \\
``\ldots my phone number is 555-123-2002.'' & unchanged & Four lookups: the customer, two lines (to find the one with this
number) and the line's plan. \\
The lookup results & unchanged & Template: check airplane mode and the SIM card. \\
``I can't perform multiple actions at once.'' & unchanged & Template: check airplane mode alone. \\
``Airplane Mode was on, so I've turned it off.'' & unknown; SIM card & Template: check the SIM card. \\
``My SIM card shows as active and working.'' & unknown; roaming & Template: are you travelling outside your home country? \\
``Yes, I'm currently abroad in France.'' & abroad, with roaming off on the line & Tool call: enable roaming on the line. Template:
turn on data roaming on the phone and try again. \\
``\ldots it still won't go through.'' & failing; data usage & Lookup: data usage (within the limit). Template: check mobile data.
\\
``Still not working. Mobile Data is on\ldots'' & failing; network mode & Template: check the network mode preference. \\
``It's working now!'' & fixed & Template: closing. \\
\bottomrule
\end{tabularx}
\end{table}

\section{The decidable share}\label{app:decidable}
We replay a frozen controller over the conversations of the LLM agent on its own. The controller receives the customer's
messages and the agent's own tool results as the agent saw them, and at each of the agent's turns it's asked, on a copy of its state, what it would do.

A turn is \emph{templatable} if the controller's state fixes its content, so that a controller could send it without a model. This is the measure we report. In telecom, it's a move from the controller's executable menu in that state. In SymPy, it's a submission or a re-run of a test the agent has already written, although our SymPy controller executes only the submission. A turn \emph{agrees} if it's exactly the controller's next move, and is \emph{anticipated} if the controller had any move ready (since a move not yet made stays proposed turn after turn, this counts loosely).

As a check, the same replay over the \blm's own conversations reproduces which turns the controller executed on all 9,227 turns of test
sets 2 and 3. For test set 1, which the first controller ran, we replay the second controller; replaying the third instead
changes no share by more than 6.4 points.

Table~\ref{tab:decidable} gives the shares. The controller's state fixes nearly all of Opus's turns in telecom, and a
small share of the SymPy agent's turns.

\begin{table}[ht]
\centering\small
\caption{Decidable share of the LLM agent's turns (in brackets, the share that agrees with the controller's next move and the
share anticipated by some move of the controller), the share of turns the \blm{} executed, and the change in model calls per
conversation. InventoryBench and CI are decision-only by construction.}
\label{tab:decidable}
\setlength{\tabcolsep}{3pt}
\begin{tabular}{@{}lccc@{}}
\toprule
Domain (baseline) & Decidable & Executed & Calls \\
\midrule
Telecom, Opus, test sets 1--3 & 98\% [55--61, 97] & 97--99\% & $-$96 to $-$99\% \\
Telecom, Qwen, test sets 1--3 & 76--80\% [26--31, 96--99] & 80--93\% & $-$74 to $-$93\% \\
SymPy, Qwen & 15--22\% [1, 15] & 3\% & $-$23\% \\
InventoryBench, CI & 100\% & 100\% & all decision calls \\
\bottomrule
\end{tabular}
\end{table}

\section{Measuring saturation}\label{app:saturation}
We replayed every conversation on a telecom training task that ran without a solved policy through the controller's frozen
reader, in the order the conversations were collected (7 to 8 October 2026, UTC). There are 1,721 such conversations on 750
distinct tasks, with 29,264 agent turns: 590 from Qwen3.8-27B on its own, 731 from Qwen3.8-27B exploration runs, and 400 from
Claude Opus 5.5 on its own. By kind of problem, 848 are picture messaging, 794 mobile data and 79 no service. In the
exploration runs an earlier controller executed the account lookups (15.6\% of their turns) and blocked some hand-offs at random. The LLM chose every other move.

As a check, we used three test runs in which the reader logged its state while the LLM agent worked on its own. They were Qwen3.8-27B on test set 3 and on unseen-fault set 2, and Qwen3.8-27B with thinking and the checklist on test set 3. On all 7,998 of their turns, the replay reproduces the logged state.

We counted four things at every agent turn. The \emph{decision state} is the kind of problem, whether it's fixed yet, anything stopping a fix, and the first check in the solved order not yet made: 69 distinct values, the last first seen at conversation 1,542. A simpler state that keeps only how many relevant checks remain (counted up to three) took 36 of its 108 possible values, the last at conversation 1,305, and that value was a misreading by the reader. It saturates partly by construction, because 96.7\% of turns fall in the
``three or more'' group.

The \emph{record of checks} adds the full set of checks made so far: 2,664 values, with new ones to
the end. The \emph{messages} are the agent's texts, lowercased, with digits, identifiers and phone numbers masked: 18,254
distinct among 22,538 text turns. Over 200 random orders of the conversations, the median number of conversations
since the last new decision state was 176 (95\% range 18 to 677), so the observed 179 is typical rather than an artifact of the order of collection.

Table~\ref{tab:coverage} gives, for each test set, how many conversations of the LLM on its own reached a decision state, or a
record of checks, not seen in training. Besides the accuracy figures of Section~\ref{sec:build}, the reader
identified anything stopping a fix in 1,716 of the 1,721 training conversations, and most of its errors on whether it was fixed read a failure as fixed.

Three caveats apply. About half of the training conversations concern mobile data, while test sets 2 and 3 and both
unseen-fault sets are all picture messaging. Saturation is measured on the reader's states, not on the true state of the phone.
And agents that behave differently from those in the training data reach unseen states more often: Qwen3.8-27B with the
checklist and thinking did so in 24 of the 114 conversations of test set 2.

\begin{table}[ht]
\centering\small
\caption{Coverage of the test sets by the training conversations: conversations of the LLM on its own that reached a decision
state, or a record of checks, not seen in training. Turns in unseen decision states in parentheses.}
\label{tab:coverage}
\begin{tabular}{@{}llccc@{}}
\toprule
Test set & Agent & Conversations & Unseen decision state & Unseen record of checks \\
\midrule
Test set 1 & Qwen alone & 114 & 0 (0 of 2,038 turns) & 50 \\
 & Opus alone & 114 & 0 (0 of 1,779) & 43 \\
Test set 2 & Qwen alone & 114 & 0 (0 of 2,095) & 65 \\
 & Opus alone & 114 & 0 (0 of 1,934) & 61 \\
Test set 3 & Qwen alone & 114 & 0 (0 of 2,380) & 62 \\
 & Opus alone & 114 & 1 (1 of 1,923) & 62 \\
Unseen-fault set 1 & Qwen alone & 60 & 0 (0 of 1,175) & 34 \\
Unseen-fault set 2 & Qwen alone & 180 & 1 (4 of 3,444) & 118 \\
 & Opus alone & 41 & 0 (0 of 670) & 23 \\
\midrule
All & & 965 & 2 (5 of 17,438) & 518 \\
\bottomrule
\end{tabular}
\end{table}

\section{Success by number of faults}\label{app:faults}
\begin{figure}[ht]
\centering
\includegraphics[width=\textwidth]{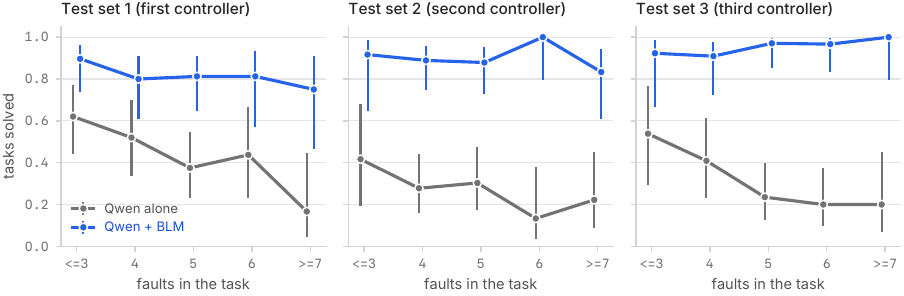}
\caption{Tasks solved by number of faults, for each test set with its own frozen controller. Points are success rates with Wilson 95\% intervals. Notice that alone, Qwen3.8-27B does worse as faults are added, and as a \blm{} it declines much less.}
\label{fig:faults}
\end{figure}

On every telecom test set, Qwen3.8-27B on its own does worse as the number of faults in a task grows, and as a Bellman model, it drops much less (Figure~\ref{fig:faults}).

\section{A simulated world for comparing decisions}\label{app:sim}
To compare an LLM's decisions with optimal ones exactly, we used a simulated world of a coding agent with hidden variables, in which the expected cost of any controller can be computed. It has 46 reachable states. An LLM given a one-page description
of the task estimated, in one pass, the probabilities of each move's outcomes. We solved the decision process over the observed state twice, once with those estimates and once with the true probabilities. This solver is a table found by dynamic programming, and it doesn't track the hidden variables. Both tables cost 29.94 minutes per task.

Asked instead to choose the move in each state, the LLM cost 32.96 minutes per task with a quick judgment and 32.21 after reasoning about what each move would reveal. Its choices agreed with the true-probability table's in 54\% and 59\% of states. A controller that knows the hidden state costs 18.82.

The solvers compare as in the real domains. Tracking a belief about the hidden variables helps: QMDP costs 27.9 minutes per task, less than the table's 29.94. Looking further ahead adds nothing (27.89 to 27.96 for horizons 1 to 6), and acting as if the most likely hidden state were certain loops (238.8). The controller that knows the hidden state is out of reach because no available move reveals the
remaining information.

The LLM roles were played by Claude subagents given only the task description. The one-pass estimates and the quick judgments came from Claude Opus 5.5, and the reasoned choices from an earlier model, Claude Opus 5, so the gap to those choices may be too large. The world, including its probabilities, was written by Claude, and its list of states was designed to be sufficient,
so this experiment shows what an LLM does with a well-posed decision problem, not how well one can be posed.

\section{CI triage in detail}\label{app:ci}
Prices come from published runner list prices (\$0.006 per minute on Linux, \$0.010 on Windows, \$0.062 on macOS) and a
developer's time at \$64.44 per hour. Take the costs of Section~\ref{sec:domains}, once no retrigger is worth its price. Marking a failure intermittent is the cheapest final move when the belief $b$ that it's a regression is below 0.125, since two developer-hours times $b$ is then less than fifteen minutes. Backing out is the cheapest above 0.75, since one developer-hour times $1-b$ is then less than fifteen minutes, and escalating is the cheapest in between.

The controller we report plans two moves ahead over $b$. Its prior comes from a logistic model of the failing test's label
history and the change that triggered it. It treats retriggers of the same failure as correlated, with a flakiness rate for each test. And the weight it gives each piece of evidence is calibrated on failures from the training window.
Table~\ref{tab:ci} also scores three variants. The first is the same controller without that calibration. The second uses the same models with a different calibration, which recalibrates the belief itself on training-window failures with observed retrigger outcomes. The third is a simpler controller with an outcome model that treats retriggers as independent and gives every test in a suite one flakiness rate.

Table~\ref{tab:ci} scores each method three ways. The first column uses only the 169 test failures with all retriggers observed, so nothing is simulated. Mozilla retriggered each of these at least twice, so they aren't a random sample, and staff
labels may reflect the retrigger results. The other two columns score all 1,173 test failures. For the 1,004 of them (86\%) with some retrigger outcomes unobserved, the missing outcomes are simulated, either from a pass rate per test fitted on all
the data or from the simpler controller's outcome model. On real outcomes, the controller we report and ``retrigger once, then escalate'' differ by $-$\$1.32 to $+$\$3.22 per
failure (95\% interval, controller minus rule).

The controller without calibration shows how a simulator can favor a controller when the two share an outcome model. It is the most expensive of the four planning controllers on real outcomes, yet the cheapest of all in the simulator that shares its per-test model.

Two further simulators share no controller's model. Both are fitted on training outcomes, one with a retrigger pass rate per label, and one with a pass rate per label and previous
outcome. With every controller fitted again on more training data, the controller without calibration was again the
cheapest (\$9.02 and \$8.47 per failure). The design we report cost \$9.40 and \$9.17, which is 25\% and 22\% less than the
best alternative, Qwen given prices and probabilities (\$12.56 and \$11.69). And the recalibrated controller was cheaper than
every rule, both Qwen controllers, QMDP and the most-likely-label solver in all four simulators (paired 95\% intervals below
zero).

Given the same prices and probabilities as the controller, Qwen3.8-27B retriggered first on 612 of 631 failures of a
development sample, and Claude Opus 5.5 on 541. Asked to read a failure's error lines, Opus was right 80.0\% of the time and
Qwen 73.4\%, against 62.0\% for always answering ``intermittent''.

\begin{table}[ht]
\centering\small
\caption{CI triage, cost per failure. Real outcomes: the 169 test failures with all retrigger outcomes observed. The
simulated columns score all 1,173 test failures, drawing unobserved retriggers from a pass rate per test fitted on all the data,
or from the simpler controller's outcome model. The best rule is chosen on the training window under each column's scoring.
The recalibrated controller was scored in a separate run. Notice that on real outcomes the simple rule and Qwen are cheapest, while in both simulated columns the planning controllers are.}
\label{tab:ci}
\setlength{\tabcolsep}{3pt}
\begin{tabular}{@{}lccc@{}}
\toprule
Policy & Real outcomes & Simulated: & Simulated: \\
 & (169 failures) & per-test rates & simpler model \\
\midrule
\blm, planning over beliefs: the controller we report & \$14.16 & \$8.70 & \$9.79 \\
\quad the same, without calibration & \$16.18 & \$7.85 & \$10.10 \\
\quad the same, recalibrated & \$14.85 & \$8.24 & \$9.51 \\
\quad the simpler controller & \$14.98 & \$10.06 & \$9.62 \\
\blm, QMDP & \$16.49 & \$12.80 & \$12.80 \\
\blm, most likely label & \$32.79 & \$20.38 & \$20.38 \\
Best re-tuned rule & \$13.60 & \$11.97 & \$13.67 \\
Retrigger once, then escalate & \textbf{\$13.23} & \$11.97 & \$13.71 \\
Qwen given prices and probabilities & \textbf{\$13.23} & \$11.82 & \$13.57 \\
\bottomrule
\end{tabular}
\end{table}

\end{document}